\documentclass{article}

\usepackage{arxiv}

\usepackage[utf8]{inputenc} 
\usepackage[T1]{fontenc}    
\usepackage{hyperref}       
\usepackage{url}            
\usepackage{booktabs}       
\usepackage{amsfonts}       
\usepackage{nicefrac}       
\usepackage{microtype}      
\usepackage{graphicx}
\usepackage{natbib}
\usepackage{doi}
\usepackage{amsmath}
\usepackage{bbm}

\title{SpikeReg: Energy-Efficient 3D Deformable Medical Image Registration with Spiking Neural Networks}

\author{
    Ali Mikaeili Barzili$^{1,2}$ \And
    Behzad Moshiri$^{1,4,*}$ \And
    Hamid Azadegan$^{3}$ \And
    Mohammad-Reza A. Dehaqani$^{1}$ \\
    \\
    $^1$ \normalfont School of Electrical and Computer Engineering, College of Engineering, University of Tehran, Tehran 1439957131, Iran \\
    $^2$ \normalfont Max Planck Institute for Brain Research, 60438 Frankfurt am Main, Germany \\
    $^3$ \normalfont School of Computer Engineering, Iran University of Science and Technology (IUST), Tehran 16846-13114, Iran \\
    $^4$ \normalfont Department of Electrical and Computer Engineering, University of Waterloo, Waterloo, Ontario N2L 3G1, Canada \\
    $^*$ \normalfont Corresponding author: \texttt{moshiri@ut.ac.ir}}

\renewcommand{\shorttitle}{SpikeReg}

\hypersetup{
pdftitle={SpikeReg: Energy-Efficient 3D Deformable Medical Image Registration with Spiking Neural Networks},
pdfsubject={cs.CV, eess.IV},
pdfauthor={Ali Mikaeili Barzili, Behzad Moshiri, Hamid Azadegan, Mohammad-Reza A. Dehaqani},
pdfkeywords={spiking neural networks, deformable image registration, brain MRI, ANN-to-SNN conversion, energy-efficient deep learning, neuromorphic computing},
}

\begin{document}
\maketitle

\begin{abstract}
	Deformable medical image registration aligns anatomical structures across
images but remains computationally dense at 3D resolution. Spiking neural
networks (SNNs) offer sparse event-driven computation, yet have not been
systematically studied for deformable medical image registration. We
introduce SpikeReg, a spiking U-Net for 3D brain MRI registration. SpikeReg
is initialized from an analog ANN registration teacher, converted by
layer-wise weight transfer and activation-percentile threshold
calibration, and fine-tuned with a surrogate-gradient objective combining
local cross-correlation, diffusion regularization, and spike-rate sparsity.
On the OASIS Learn2Reg validation split ($19$ image pairs), SpikeReg reaches
Dice $0.7474 \pm 0.032$, with no significant paired Dice difference from the ANN
teacher ($0.7480 \pm 0.037$, $p = 0.67$), at a $12.8\%$ mean spike rate and
a $55.5\times$ projected arithmetic-energy reduction under an event-sparse
SynOps/MAC proxy relative to the dense-ANN baseline. We additionally report two negative findings:
displacement distillation from the ANN teacher hurts performance, and ANN
teachers trained with a label-Dice loss fail to transfer through rate-code
conversion. Together these results show that dense geometric prediction can
be performed under sparse event-driven computation, opening a path toward
neuromorphic medical image registration.

\end{abstract}


\section{Introduction}
\label{sec:introduction}


Deformable medical image registration aims to estimate a spatial transformation that aligns anatomical structures across images. In brain MRI this operation underpins atlas propagation, longitudinal monitoring, and population-level morphometry. Unlike image classification, where a network predicts one or a small number of global labels, deformable registration requires a dense three-dimensional transformation field assigning every voxel a spatial mapping. The registration model must therefore preserve local anatomical detail while also capturing long-range spatial relationships between the moving and fixed images.

Classical registration algorithms cast this problem as an optimization over deformation fields, typically combining an image-similarity term with spatial regularization. Diffeomorphic methods such as SyN and diffeomorphic Demons further constrain the transformation to be smooth and invertible, treating topology preservation as a primary registration constraint rather than a secondary evaluation metric \citep{avants2008syn,vercauteren2009diffeomorphic}. Their iterative optimization can be expensive when applied across large 3D cohorts, making amortized inference attractive: train the network once and reuse it across the full cohort without re-solving the optimization per pair.


Deep learning has substantially changed this landscape. VoxelMorph showed that convolutional networks can perform unsupervised pairwise deformable registration with accuracy comparable to classical methods while reducing inference time by orders of magnitude \citep{balakrishnan2019voxelmorph}. Probabilistic and diffeomorphic variants added uncertainty modelling and stationary velocity-field integration to improve transformation regularity \citep{dalca2019probabilistic}, and transformer-based models such as TransMorph introduced long-range self-attention to volumetric registration, establishing strong performance on brain MRI benchmarks including OASIS/Learn2Reg \citep{chen2022transmorph}.

Despite these advances, the dominant learning-based registration paradigm remains computationally dense. CNN- and transformer-based registration networks evaluate convolution, attention, normalization, and interpolation over the full volume, regardless of whether all spatial locations contribute equally to the final deformation. Memory and arithmetic cost scale cubically with the spatial extent of the volume, and current benchmarks emphasize Dice, Hausdorff distance, target registration error, Jacobian statistics, and runtime \citep{hering2023learn2reg}. Explicit modelling of sparse computation and energy-aware inference is comparatively underexplored in deformable registration.


Spiking neural networks (SNNs) provide a different computational substrate. Instead of propagating dense real-valued activations at every layer, SNNs communicate through sparse binary events over time. This event-driven representation is attractive for low-power inference because inactive neurons skip unnecessary synaptic operations, especially on neuromorphic hardware designed for sparse asynchronous computation \citep{davies2018loihi,davies2021loihi}. In conventional artificial neural networks, convolutional layers are dominated by multiply--accumulate operations; in spiking networks, many synaptic updates reduce to conditional accumulate operations triggered only by spikes, making the effective cost a function of measured spike activity rather than network size alone \citep{horowitz2014computing,rueckauer2017conversion}.

However, applying SNNs to deformable image registration is not a direct extension of existing SNN applications. Most successful ANN-to-SNN conversion work targets image classification, where the prediction is low-dimensional and errors are averaged over output classes \citep{diehl2015fast,rueckauer2017conversion}. Medical-image SNNs have begun to address segmentation and neuroimaging tasks: U-Net-like SNNs for brain MRI \citep{patel2021spikingseg,yue2023snnseg}, and the recent Spiking-UNet ANN-to-SNN conversion plus surrogate-fine-tune pipeline for 2D images \citep{li2024spikingunet}. Dense deformable registration is more demanding: the output is continuous, high-dimensional, spatially coupled, and directly tied to deformation topology. A small local error in the displacement field can reduce anatomical overlap, introduce unrealistic local distortion, or produce non-positive Jacobian determinants. The central question is therefore not whether a spiking network can process medical images, but whether sparse temporal spikes can support accurate and smooth \emph{dense geometric regression}.


This gap motivates the present work. To our knowledge, no prior work systematically examines spiking neural networks for 3D deformable medical image registration under a registration-specific evaluation protocol. Existing registration methods provide strong accuracy and topology baselines, while existing SNN methods provide conversion, surrogate-gradient training, and energy-efficient inference. What is missing is a framework that connects the two: a spiking model that predicts dense 3D deformation fields, is trained with registration losses, is warm-started from an ANN registration teacher, and is evaluated not only by anatomical alignment but also by deformation regularity and spike-dependent computational cost.


We introduce \textit{SpikeReg}, a spiking neural framework for energy-aware deformable medical image registration. SpikeReg follows an ANN-to-SNN training strategy tailored to dense geometric prediction. First, an ANN U-Net registration model is trained to learn stable spatial features and displacement prediction. The ANN is then converted into a spiking U-Net by transferring convolutional and normalization parameters and calibrating layer-wise firing thresholds from ANN activation statistics. The resulting SNN is fine-tuned with surrogate gradients, allowing the network to adapt its temporal spike dynamics to the registration objective. During training, continuous image intensities are injected as input current rather than stochastic Poisson spike trains, preserving gradient stability while allowing spike-based computation in the hidden layers. The model outputs a dense three-channel displacement field and records layer-wise spike activity, supporting direct analysis of the relationship between spike sparsity, registration accuracy, and estimated arithmetic energy.

The training objective combines image similarity, deformation regularization, spike regularization, and optional teacher distillation. For mono-modal brain MRI registration, local normalized cross-correlation is used as the image similarity term, while diffusion regularization encourages smooth displacement fields. Spike-rate regularization discourages excessive firing while avoiding silent layers, and distillation can in principle be added as a tunable constraint toward the ANN teacher after conversion. This design reflects a key property of the task: in deformable registration, energy reduction cannot be pursued independently of deformation quality. Excessive sparsity may under-drive the displacement field, whereas uncontrolled firing erodes the computational advantage of the SNN. SpikeReg therefore treats spike activity as an explicit part of the registration trade-off rather than as an implementation detail.


We evaluate SpikeReg on the OASIS/Learn2Reg inter-subject brain MRI registration task. Primary anatomical accuracy is mean label-wise Dice after warping the moving segmentation; surface accuracy is HD95; image alignment is NCC; deformation regularity is the fraction of voxels with non-positive Jacobian determinant. In addition, we report spike activity, projected SynOps (from hook-measured spike counts and ANN MACs), multiply--accumulate baselines, and analytical energy proxies derived from measured firing rates. The evaluation is deliberately not framed as a leaderboard chase: the goal is to characterize whether dense 3D registration can be performed under sparse event-driven computation while retaining most of the accuracy of its ANN counterpart.


Our contributions are fourfold. \emph{First}, we formulate 3D deformable medical image registration as a dense spiking regression problem and present, to our knowledge, the first systematic framework for SNN-based 3D deformable registration. \emph{Second}, we adapt the standard ANN-to-SNN conversion plus surrogate-gradient fine-tuning pipeline \citep{rueckauer2017conversion,li2024spikingunet} to dense geometric regression, and empirically characterize which design choices transfer (ANN warm-start, activation-percentile threshold calibration) and which do not. The latter are reported as standalone negative findings: displacement-MSE distillation over-constrains voxel-scale outputs, label-supervised ANN teachers fail to transfer through rate-code conversion, increasing the temporal budget to $T=6$ remains unstable even after fresh calibration and seed replication, and a stationary-velocity-field SpikeReg variant improves topology at the cost of an unacceptable Dice loss. \emph{Third}, we characterize the accuracy--energy trade-off at the temporal budgets we completed ($T \in \{2,4,6\}$, plus raw-conversion threshold sweeps at $T=4$) and identify $T=4$ with percentile $p=50$ as the dominant operating point in this sweep, at which the SNN matches its ANN teacher within statistical noise; expanded sweeps over $T \in \{1,8,12\}$ remain future work. \emph{Fourth}, we provide an accuracy--topology--energy evaluation protocol that reports registration metrics together with per-layer spike rates, projected synaptic operation counts from hook-measured spike activity, and an analytical arithmetic-energy proxy. Together these contributions move dense registration toward energy-aware geometric modelling and lay groundwork for neuromorphic deployment.

\section{Related Work}
\label{sec:related_work}

\paragraph{Classical and learning-based registration.}
Classical deformable registration solves a per-pair optimization that
balances image similarity against deformation regularity. Free-form
deformation models with B-spline parameterizations
\citep{rueckert1999nonrigid}, symmetric normalization (SyN)
\citep{avants2008syn}, and diffeomorphic Demons
\citep{vercauteren2009diffeomorphic} remain widely used as
non-learning baselines, particularly in volumetric neuroimaging, because
they explicitly model topology preservation through diffeomorphic
constraints. Learning-based registration replaces this per-pair
optimization with a single trained network: VoxelMorph
\citep{balakrishnan2019voxelmorph} predicts dense displacement fields
under unsupervised similarity and smoothness losses; probabilistic and
diffeomorphic variants integrate stationary velocity fields for
topology-preserving inference \citep{dalca2018diffeomorphic,dalca2019probabilistic};
recursive-cascaded networks \citep{zhao2019recursive} and Laplacian-pyramid
methods \citep{mok2020lapirn} address large deformations through progressive
warping; and transformer-based models such as TransMorph
\citep{chen2022transmorph} introduce long-range attention to volumetric
correspondence. These methods define the methodological baseline against
which SpikeReg is positioned: SpikeReg adopts the same U-Net macro-architecture
and unsupervised similarity-plus-regularization training strategy, but
replaces the dense activation pathway with sparse temporal spiking
computation. SpikeReg is therefore not in competition with these networks
on absolute Dice; the relevant axis is whether the registration function
they learn survives a transition to event-driven inference.

\paragraph{Training spiking neural networks.}
Two complementary families of SNN training methods are directly relevant.
Surrogate-gradient learning treats the non-differentiable spike function
through a smooth approximation only on the backward pass, enabling
end-to-end training through deep spiking networks
\citep{zenke2018superspike,neftci2019surrogate}. ANN-to-SNN conversion
takes a pretrained analog network and replaces ReLU activations with
spiking neurons whose firing rate approximates the analog activation;
threshold balancing and per-layer activation-percentile calibration are
the central mechanisms for preserving accuracy after conversion
\citep{diehl2015fast,rueckauer2017conversion}. Residual membrane
potential neurons extend this recipe by carrying suprathreshold charge
into the next timestep rather than discarding it on reset, recovering
much of the conversion loss that hard-reset neurons incur at low temporal
budgets \citep{han2020rmp}. The Spiking-UNet line of
work \citep{li2024spikingunet} shows that the warm-start, conversion, and
surrogate-fine-tune recipe transfers from classification to 2D U-Net
image-level prediction; however, these demonstrations assume discrete or
categorical targets, and conversion errors have been shown to compound
when the output is a continuous signal---small approximation errors
become temporally correlated and induce cumulative distribution shift in
continuous control settings \citep{xu2026erroramp}. SpikeReg adopts this same recipe but applies it
to a fundamentally different output target: a continuous,
spatially-coupled 3D displacement field rather than a per-pixel class
distribution.

\paragraph{SNNs for dense prediction in medical imaging.}
Most successful deep SNN demonstrations target classification or
event-based vision; dense medical prediction is comparatively rare.
Recent work has converted U-Net-like architectures to spiking
computation for medical and biomedical \emph{segmentation}
\citep{patel2021spikingseg,yue2023snnseg}, including analog-trained
variants for natural-image segmentation that have been adapted to medical
data \citep{ma2025analog}. Earlier work also used spiking-cortical-model
similarity features inside otherwise classical multi-modal registration
pipelines \citep{zhu2017scmregistration}, but did not train a deep
spiking encoder--decoder to predict a dense deformation field.
Parallel efforts in scientific machine learning have shown that spiking
networks can approximate continuous functions and operator mappings with
substantial energy reductions \citep{kahana2022spiking}, but those
demonstrations operate on low-dimensional function spaces rather than
spatially coupled volumetric fields with geometric constraints. Segmentation
predicts discrete labels at each voxel; registration predicts a continuous
3D vector field whose spatial derivatives determine local volume change,
folding, and invertibility. A sparse spiking representation of a
displacement field must therefore preserve sub-voxel magnitudes across
homogeneous anatomy, not only at label boundaries. This is a stricter test of
SNN expressivity than anything reported for medical SNNs to date.

\paragraph{Energy-aware reporting and the gap addressed by SpikeReg.}
Modern registration networks are fast at inference but computationally
dense: convolutional and transformer registration models evaluate dense
multiply--accumulate operations across the full volume regardless of
local relevance. SNNs offer a different regime in principle, but
defensible energy claims must distinguish between hardware-measured
energy on neuromorphic devices \citep{davies2018loihi,davies2021loihi}
and architecture-level analytical operation-count proxies that depend on
measured spike rates and per-layer fan-in
\citep{rueckauer2017conversion,lemaire2022analytical}. Registration
benchmarks such as Learn2Reg \citep{hering2023learn2reg} score Dice,
HD95, and Jacobian regularity but do not currently include a
sparsity- or operation-count axis. Within this landscape, SpikeReg
contributes, to our knowledge, the first systematic study of
deep-spiking dense 3D deformable registration, positioned not as a
Dice-leaderboard entry but as a joint accuracy--topology--energy
characterization of what survives ANN-to-SNN conversion for a continuous
geometric prediction task.

\section{Method}
\label{sec:method}

\begin{figure}[htbp]
\centering
\includegraphics[width=\linewidth]{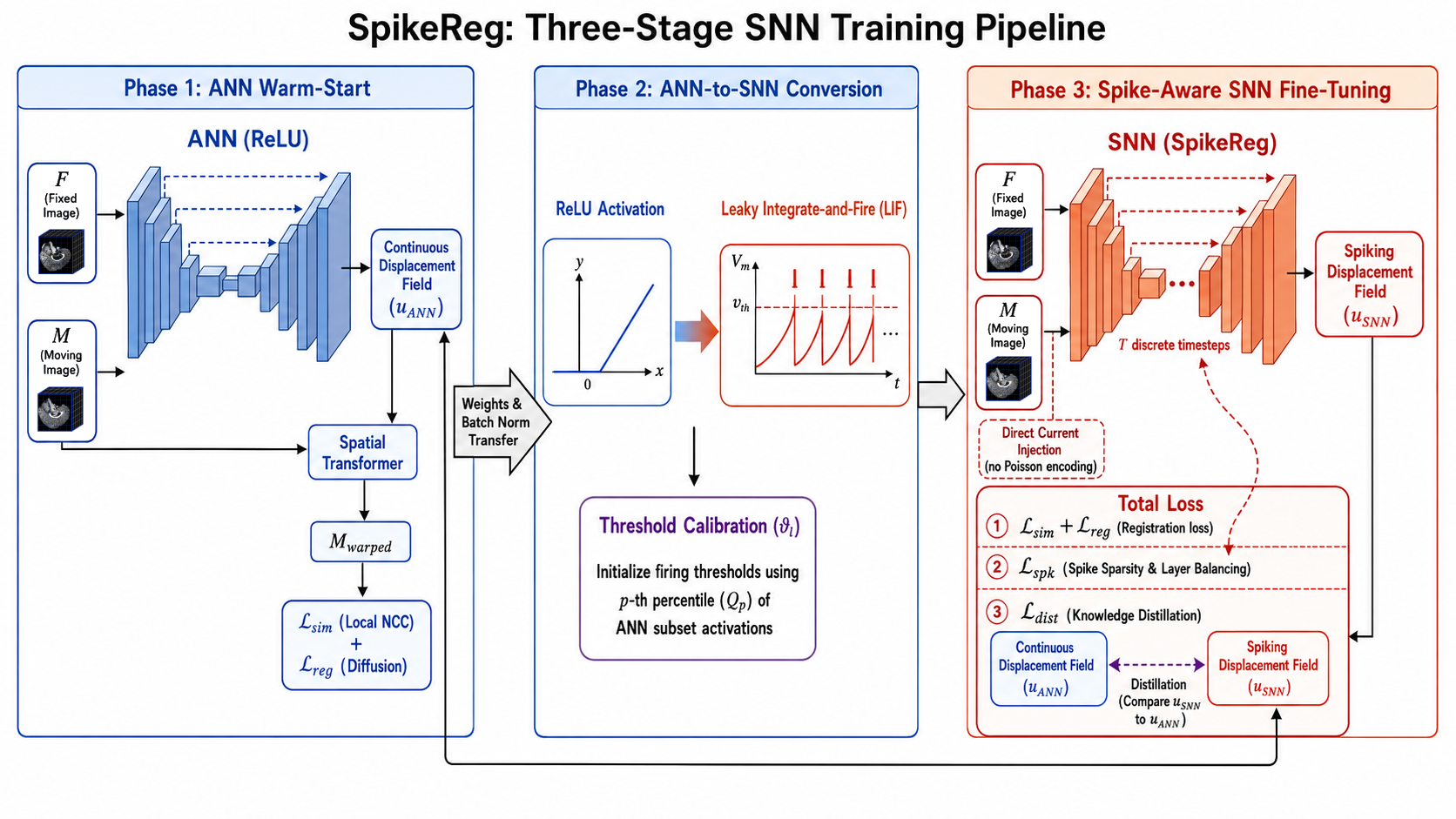}
\caption{
SpikeReg three-stage training pipeline.
An analog 3D U-Net is first trained as the registration teacher with $\mathcal{L}_{\mathrm{sim}} + \mathcal{L}_{\mathrm{reg}}$ (Phase~1). Convolutional and batch-normalization parameters are then transferred layer-by-layer to a spiking U-Net, and per-layer firing thresholds are calibrated from the $50$-th activation percentile on a held-out subset (Phase~2). The converted SNN is fine-tuned with surrogate gradients under the canonical objective in Eq.~\eqref{eq:canonical_loss} (Phase~3); the dashed displacement-distillation term $\mathcal{L}_{\mathrm{dist}}$ shown in Phase~3 of the figure is evaluated as an ablation variant only and is \emph{not} part of the canonical loss (see Section~\ref{sec:results}).
}
\label{fig:pipeline}
\end{figure}

We propose \emph{SpikeReg}, a spiking neural network framework for
unsupervised 3D deformable medical image registration. Given a fixed image
$F:\Omega \rightarrow \mathbb{R}$ and a moving image
$M:\Omega \rightarrow \mathbb{R}$ on a 3D voxel lattice
$\Omega \subset \mathbb{R}^{3}$, the goal is to estimate a dense
transformation $\phi:\Omega \rightarrow \Omega$ that aligns $M$ to $F$.
In the direct-displacement formulation, the network predicts a voxel-wise
displacement field $u:\Omega \rightarrow \mathbb{R}^{3}$, and the
transformation is simply
\begin{equation}
    \phi(x) = x + u(x).
\end{equation} The warped moving image is then
$M_{\phi}(x) = M(\phi(x))$, obtained by differentiable trilinear sampling
\citep{jaderberg2015spatial}; out-of-volume coordinates are clamped to the
nearest boundary voxel, with grid coordinates aligned to voxel centers. Unlike standard convolutional registration
networks, SpikeReg replaces the dense activation pathway with temporally
sparse spiking computation while preserving the U-Net encoder--decoder
structure required for dense 3D prediction. The full pipeline is summarized
in Figure~\ref{fig:pipeline}.

\paragraph{Network architecture.}
SpikeReg takes the fixed and moving images as a stacked two-channel input
$X = [F, M] \in \mathbb{R}^{2 \times D \times H \times W}$ and produces a
three-channel deformation field
$\hat{u} = G_{\theta}^{\mathrm{SNN}}(F,M) \in \mathbb{R}^{3 \times D \times H \times W}$.
The backbone is a 3D U-Net with four encoder levels, a bottleneck, a
symmetric decoder, and skip connections: encoder channel widths
$[16,32,64,128]$, decoder widths $[64,32,16,16]$, matching the
VoxelMorph-style macro-architecture \citep{balakrishnan2019voxelmorph}.
The encoder halves spatial resolution at each level while doubling channels;
the decoder mirrors this, pulling in skip features to recover the fine-grained
spatial detail that displacement prediction requires. The final projection
head collapses the decoder spike sequence into a continuous three-channel
field.

\paragraph{Spiking convolutional block and LIF dynamics.}
Each encoder and decoder block replaces the ReLU nonlinearity with a leaky
integrate-and-fire (LIF) neuron. Let $I_{\ell,t}$ denote the pre-integrated
input current to layer $\ell$ at timestep $t$ (the output of the preceding
3D convolution and batch normalization). The membrane potential evolves as
\begin{equation}
    v_{\ell,t} = \tau_\ell\, v_{\ell,t-1} + I_{\ell,t},
\end{equation}
and a binary spike fires whenever the membrane crosses the layer's calibrated
threshold $\vartheta_\ell$:
\begin{equation}
    s_{\ell,t} = H(v_{\ell,t} - \vartheta_\ell),
\end{equation}
after which the membrane resets to zero via $v_{\ell,t} \leftarrow v_{\ell,t}(1 - s_{\ell,t})$.
The leak $\tau_\ell$ is initialized to a symmetric U-shaped schedule: it decreases from
$0.90$ to $0.75$ through the encoder (shallow to deep) and rises back from
$0.75$ to $0.90$ through the decoder. Because $v_{\ell,t} = \tau_\ell v_{\ell,t-1} + I_{\ell,t}$,
a higher $\tau$ means slower decay and longer temporal memory.
Shallower layers (encoder $\tau = 0.90$, decoder $\tau = 0.90$) therefore retain a longer
temporal memory, allowing them to accumulate fine-grained intensity differences across the $T$
timesteps---useful for the sub-voxel precision that high-resolution displacement prediction requires.
The deeper encoder stages and bottleneck use a shorter memory ($\tau = 0.75$), keeping each
timestep's abstract representation responsive to the current input rather than dominated by
accumulated state. During surrogate fine-tuning these per-channel leak coefficients are
updated as learnable parameters, so the schedule serves as an initialization rather than a
fixed constraint.

Because the Heaviside step is non-differentiable, we use a fast-sigmoid
surrogate during backpropagation through time
\citep{zenke2018superspike,neftci2019surrogate}. On the backward pass, the
spike is treated as $\sigma(\alpha(v_{\ell,t} - \vartheta_\ell))$, giving
the gradient approximation
\begin{equation}
    \frac{\partial s_{\ell,t}}{\partial v_{\ell,t}}
    \approx
    \alpha\,\sigma\!\bigl(\alpha(v_{\ell,t} - \vartheta_\ell)\bigr)
    \bigl(1 - \sigma\!\bigl(\alpha(v_{\ell,t} - \vartheta_\ell)\bigr)\bigr),
\end{equation}
with steepness $\alpha = 10$. The forward pass remains binary; the backward-pass
surrogate is evaluated in single precision even when the surrounding network
operates under 16-bit mixed-precision training, since at $\alpha = 10$ the
reduced dynamic range of 16-bit formats clips gradients for membrane values
closest to the threshold, which carry the most learning signal.

\paragraph{Spike-to-rate interface.}
Because the output must be a continuous displacement field, SpikeReg converts
spike trains to rate features between blocks. For layer $\ell$ over $T_\ell$
timesteps,
\begin{equation}
    r_{\ell} = \frac{1}{T_{\ell}} \sum_{t=1}^{T_{\ell}} s_{\ell,t}.
\end{equation}
These rate tensors are passed as input to the next block and along the skip
pathways; the final projection head averages the decoder spike sequence in
the same way to recover a real-valued field. Using rates at block boundaries
keeps the model compatible with standard convolutional arithmetic and lets
spike activity be measured and regularized per layer independently.
The PyTorch reference implementation passes mean-rate tensors between blocks
for gradient flow under surrogate training; the downstream convolutions are
evaluated densely. Spike counts at block outputs are recorded via forward hooks
and used to project the cost of an equivalent event-driven implementation
(Section~\ref{sec:experiments}).

For input encoding, we inject image intensities as a direct current rather
than drawing Poisson spike trains. This avoids adding stochastic sampling
noise to the dense regression objective, which would inflate gradient variance
and destabilize fine-tuning. The trade-off is that layer~1 performs
multiply--accumulate rather than spike-triggered accumulate operations; we
account for this in both the projected and analytical energy estimates so
the reported SynOps figure is not under-counted.

\paragraph{ANN warm-start.}
Training a spiking network directly for dense 3D displacement regression is
unstable: the task requires smooth, sub-voxel-precise outputs at every spatial
location, and surrogate gradients alone cannot reliably bootstrap this from a
random initialization. We therefore first train $G_{\psi}^{\mathrm{ANN}}$, an analog U-Net with the
same macro-architecture but ReLU activations in place of LIF neurons, under
the unsupervised registration objective
\begin{equation}
    \mathcal{L}_{\mathrm{ANN}}
    =
    \lambda_{\mathrm{sim}}\mathcal{L}_{\mathrm{sim}}(F,M_{\phi})
    +
    \lambda_{\mathrm{reg}}\mathcal{L}_{\mathrm{reg}}(\hat{u}^{\mathrm{ANN}}).
\end{equation}
The ANN learns stable spatial correspondences and smooth displacement patterns
grounded in the registration objective, and these weights are then transferred
into the spiking model rather than rediscovered through surrogate gradients
from scratch.

\paragraph{ANN-to-SNN conversion and threshold calibration.}
After ANN pretraining, convolutional and batch-normalization parameters are
copied layer-by-layer into the corresponding spiking encoder, bottleneck, and
decoder blocks. The conversion principle is that each LIF neuron's firing rate
should approximate the analog activation of its ReLU counterpart
\citep{diehl2015fast,rueckauer2017conversion}, the same recipe used by
Spiking-UNet for 2D image prediction \citep{li2024spikingunet}. The critical
free parameter is the per-layer firing threshold $\vartheta_\ell$: too low and
neurons fire densely, losing sparsity without improving accuracy; too high and
layers go nearly silent and under-drive the displacement field. Letting $\mathcal{A}_\ell$ denote the positive ANN activations at layer $\ell$
recorded on $50$ held-out pairs, we set each threshold to their $p$-th
percentile $Q_p$:
\begin{equation}
    \vartheta_\ell = Q_p\!\left(\{a \in \mathcal{A}_\ell : a > 0\}\right).
\end{equation} The
canonical model uses $p = 50$. During subsequent surrogate fine-tuning,
batch-normalization statistics and affine parameters $(\gamma,\beta)$ are
frozen to preserve the ANN feature statistics on which threshold calibration
is based. Convolutional weights, per-channel leak coefficients $\tau_\ell$,
and firing thresholds $\vartheta_\ell$ are all updated; the percentile-calibrated
thresholds and the U-shaped leak schedule serve as initializations rather than
fixed constraints, contributing the $+10{,}935$ learnable parameters that
distinguish the SNN from the ANN teacher (Table~\ref{tab:main_results}).

\paragraph{Registration losses.}
For mono-modal brain MRI we use local normalized cross-correlation (NCC) as
the image similarity term. Letting $\bar{F}_{x}$ and $\bar{M}_{\phi,x}$ be
the intensity means over a $9^3$ window $\mathcal{W}_x$ centered at $x$, and
$\epsilon = 10^{-8}$ a denominator stabilizer, the local NCC is
\begin{equation}
    \mathrm{NCC}_{x}(F,M_{\phi})
    =
    \frac{
    \sum_{y \in \mathcal{W}_{x}}
    \left(F(y)-\bar{F}_{x}\right)
    \left(M_{\phi}(y)-\bar{M}_{\phi,x}\right)
    }{
    \sqrt{
    \sum_{y \in \mathcal{W}_{x}}
    \left(F(y)-\bar{F}_{x}\right)^2
    }
    \sqrt{
    \sum_{y \in \mathcal{W}_{x}}
    \left(M_{\phi}(y)-\bar{M}_{\phi,x}\right)^2
    }
    + \epsilon
    },
\end{equation}
Averaged over the volume, the similarity loss is
\begin{equation}
    \mathcal{L}_{\mathrm{sim}}(F,M_{\phi})
    =
    -
    \frac{1}{|\Omega|}
    \sum_{x \in \Omega}
    \mathrm{NCC}_{x}(F,M_{\phi}).
\end{equation}
NCC is preferred over mean-squared error because inter-subject brain MRI scans
can differ in absolute intensity for reasons unrelated to anatomy; MSE would
penalize these as geometric misalignment, whereas NCC is invariant to local
affine intensity relationships. Deformation smoothness is encouraged by
diffusion regularization on the predicted displacement:
\begin{equation}
    \mathcal{L}_{\mathrm{reg}}(u)
    =
    \frac{1}{|\Omega|}
    \sum_{x \in \Omega}
    \|\nabla u(x)\|_{2}^{2},
\end{equation}
which penalizes sharp local spatial gradients without constraining the field's
global magnitude or direction. This makes it a natural prior for inter-subject
brain MRI, where large deformations are expected and their spatial scale is
not known in advance.

\paragraph{Spike-aware fine-tuning objective.}
During SNN fine-tuning the registration losses are augmented with a spike-rate
term to explicitly manage the accuracy--efficiency trade-off. The canonical
objective is
\begin{equation}
\label{eq:canonical_loss}
    \mathcal{L}_{\mathrm{SNN}}
    =
    \lambda_{\mathrm{sim}}\,\mathcal{L}_{\mathrm{sim}}(F, M_\phi)
    +
    \lambda_{\mathrm{reg}}\,\mathcal{L}_{\mathrm{reg}}(\hat{u}^{\mathrm{SNN}})
    +
    \lambda_{\mathrm{spk}}\,\mathcal{L}_{\mathrm{spk}}.
\end{equation}
The spike regularizer addresses a tension specific to spiking registration:
too little firing under-drives the displacement field; too much erases the
computational advantage. Let
$\rho_\ell = (T N_\ell)^{-1}\sum_{t=1}^{T}\sum_{i=1}^{N_\ell} s_{\ell,t,i}$
be the mean spike rate of layer $\ell$ over $N_\ell$ neurons and $T$
timesteps. Targeting $\rho^\star = 0.1$ spikes per timestep per neuron, we define
\begin{equation}
    \mathcal{L}_{\mathrm{spk}}
    =
    \sum_\ell \rho_\ell
    \;+\;
    \beta \sum_\ell \bigl(\rho_\ell - \rho^\star\bigr)^{2}.
\end{equation}
The first term applies a global
downward pressure on firing across all layers; the second penalizes layers
that deviate far from $\rho^\star$ in either direction, preventing the
degenerate regime where one layer fires densely while another goes silent.
Canonical weights are $\lambda_{\mathrm{sim}} = 1.0$,
$\lambda_{\mathrm{reg}} = 0.1$, $\lambda_{\mathrm{spk}} = 10^{-4}$,
$\beta = 10^{-2}$.

\paragraph{Loss variants used in ablations.}
Two additional loss terms are evaluated as ablation variants and are
\emph{not} part of the canonical objective in Eq.~\eqref{eq:canonical_loss}.

\emph{Displacement distillation} (KD) penalizes mean-squared deviation
between the SNN and the frozen ANN teacher displacement fields,
\begin{equation}
    \mathcal{L}_{\mathrm{distill}}
    =
    \bigl\| \hat{u}^{\mathrm{SNN}} - \hat{u}^{\mathrm{ANN}} \bigr\|_2^{2},
\end{equation}
weighted by $\lambda_{\mathrm{distill}}$. The motivation is that spike-induced
local errors can accumulate spatially across a dense field, and anchoring the
SNN near the teacher might stabilize fine-tuning. As reported in
Section~\ref{sec:results}, the canonical no-KD model outperforms KD; the
mechanism (voxel-scale MSE overwhelming the NCC gradient) is taken up in the
Discussion.

\emph{Soft label-Dice} for semi-supervised experiments with anatomical labels:
\begin{equation}
    \mathcal{L}_{\mathrm{segDice}}
    =
    1
    -
    \frac{1}{C}
    \sum_{c=1}^{C}
    \frac{
        2 \sum_{x \in \Omega} S^F_c(x)\, S^M_c(\phi(x)) + \epsilon
    }{
        \sum_{x \in \Omega} S^F_c(x)
        +
        \sum_{x \in \Omega} S^M_c(\phi(x))
        +
        \epsilon
    },
\end{equation}
weighted by $\lambda_{\mathrm{seg}}$. Moving segmentations are one-hot
encoded and warped by trilinear interpolation to keep the term differentiable
with respect to $\phi$; nearest-neighbor warping is used at evaluation time
only. This term is used in the ANN-teacher sweep that yields a higher-Dice
but non-transferable teacher, reported as a negative result in the Discussion.

\paragraph{Stationary-velocity-field diagnostic variant.}
To test whether explicit topology constraints are compatible with the spiking
pipeline, we evaluate a SpikeReg-SVF variant in which the network predicts a
stationary velocity field $v(x)$ and the deformation is recovered by
scaling-and-squaring integration of $\exp(v)$. The architecture, warm-start, conversion, and fine-tuning protocol are
otherwise unchanged. This variant is not the
canonical method: as reported in Section~\ref{sec:results}, it trades a large
reduction in folding for an unacceptable Dice loss, indicating that SVF
integration cannot simply be dropped in as a topology fix without redesigning
the training objective and fine-tuning schedule.

\paragraph{Arithmetic-energy proxy.}
To quantify the efficiency gain from sparse spiking activity, we use a
per-layer MAC/AC proxy corroborated by direct operation counts from the forward pass. In the
ANN, every convolutional layer evaluates a full set of multiply--accumulate
(MAC) operations, giving total energy $E_{\mathrm{ANN}} = e_{\mathrm{MAC}}\sum_\ell \mathrm{MAC}_\ell$.
In the SNN, only active presynaptic spikes trigger accumulation, so the
operation count scales with measured firing rate. Using per-layer mean rate
$\rho_\ell$ and timestep count $T$:
\begin{equation}
    E_{\mathrm{SNN}}
    =
    e_{\mathrm{AC}} \sum_\ell T\,\rho_\ell\,\mathrm{MAC}_\ell,
    \qquad
    R_{\mathrm{energy}}
    =
    \frac{E_{\mathrm{SNN}}}{E_{\mathrm{ANN}}}
    =
    \frac{e_{\mathrm{AC}}}{e_{\mathrm{MAC}}}
    \cdot
    \frac{\sum_\ell T\,\rho_\ell\,\mathrm{MAC}_\ell}{\sum_\ell \mathrm{MAC}_\ell},
\end{equation}
using $e_{\mathrm{AC}} = 0.9\,\mathrm{pJ}$ and
$e_{\mathrm{MAC}} = 4.6\,\mathrm{pJ}$ \citep{horowitz2014computing}. We report
two instantiations. The \emph{per-layer projected} version measures ANN MACs
directly from dense PyTorch forward-pass hooks, and projects SNN cost by
multiplying hook-recorded output spike counts at each block by the downstream
convolutional fan-in to obtain SynOps---the cost those spikes would incur on
an event-driven implementation. The \emph{uniform-rate analytical} version
replaces each $\rho_\ell$ with the network-wide mean $\bar\rho$, giving a
coarser upper bound convenient for comparison with conversion literature.
In our network the highest-MAC decoder layer fires at $10.7\%$, below
$\bar\rho = 12.8\%$, so the uniform approximation over-counts SNN operations;
the per-layer projected figure is therefore our headline number.
Neither value is a hardware energy measurement: the PyTorch reference
implementation executes dense convolutions and does not itself realize this
reduction; absolute claims require deployment on neuromorphic hardware
\citep{davies2018loihi,indiveri2015memory,rueckauer2017conversion}.

\paragraph{Training protocol.}
Training proceeds in two stages, whose hyperparameters are listed in
Table~\ref{tab:hparams}.

\begin{table}[htbp]
\centering
\caption{Hyperparameters for the ANN warm-start and SNN fine-tuning stages.}
\label{tab:hparams}
\begin{tabular}{lcc}
\toprule
& \textbf{ANN stage} & \textbf{SNN stage} \\
\midrule
Epochs                          & 300           & 200 \\
Optimizer                       & Adam          & Adam \\
$(\beta_1, \beta_2, \epsilon)$  & $(0.9,\;0.999,\;10^{-8})$ & $(0.9,\;0.999,\;10^{-8})$ \\
Learning rate                   & $10^{-4}$     & $10^{-4}$ \\
$\eta_{\min}$ (cosine decay)    & $10^{-6}$     & $10^{-5}$ \\
Batch size                      & 1             & 1 \\
Gradient clipping ($\ell_2$)    & 1.0           & 1.0 \\
NCC window                      & $9^3$         & $9^3$ \\
$\lambda_{\mathrm{sim}}$        & 1.0           & 1.0 \\
$\lambda_{\mathrm{reg}}$        & 0.1           & 0.1 \\
$\lambda_{\mathrm{spk}}$        & --            & $10^{-4}$ \\
$\beta$ (rate balancing)        & --            & $10^{-2}$ \\
$\rho^\star$                    & --            & 0.1 \\
$T$ (timesteps)                 & --            & 4 \\
Threshold percentile $p$        & --            & 50 \\
Calibration pairs               & --            & 50 \\
Mixed precision                 & 16-bit + single-prec.\ surrogate & 16-bit + single-prec.\ surrogate \\
Random seed                     & 42            & 42 \\
\bottomrule
\end{tabular}
\end{table}

Both stages train at full $160 \times 192 \times 224$ resolution on
$4\times$ AMD Instinct MI300A APUs ($128$\,GB unified HBM3 per device)
with data-parallel training across four devices and 16-bit mixed-precision
throughout, except for the surrogate gradient, which is kept in single
precision for the reasons described above. ANN warm-start required
approximately $2$\,h~$20$\,min wall-clock; SNN fine-tuning required
approximately $16$\,h~$15$\,min. Input preprocessing follows the publicly
released dataset from \citet{chen2022transmorph} at $160 \times 192 \times 224$
resolution.

\paragraph{Deformation regularity metrics.}
In addition to anatomical overlap, SpikeReg is evaluated as a geometric model
whose output must remain spatially coherent. Deformation regularity is
quantified by the fraction of voxels with non-positive Jacobian determinant,
\begin{equation}
    \mathrm{Fold}(\phi)
    =
    \frac{1}{|\Omega|}
    \sum_{x \in \Omega}
    \mathbbm{1}
    \bigl[
    \det\nabla\phi(x) \leq 0
    \bigr],
\end{equation}
and by the standard deviation of $\log\det\nabla\phi$ (SDlogJ), which captures
local regularity deviations even where outright folding has not yet appeared.
For each inference pair, SpikeReg records the deformation field, warped moving
image, and layer-wise spike rates and counts, enabling joint evaluation of
registration quality, deformation regularity, and event sparsity in a single
forward pass.

\section{Experiments}
\label{sec:experiments}

This section describes the experimental protocol used to evaluate SpikeReg.
All registration numbers are reported in Section~\ref{sec:results}; here we
specify the dataset, models, training set-up, evaluation metrics, and
statistical procedures.

\paragraph{Objective.}
The experiments evaluate whether dense 3D deformable registration can be
performed under sparse spiking computation while retaining most of the
accuracy of an equivalent artificial neural network (ANN) registration
model. We therefore assess SpikeReg across three dimensions: registration
accuracy, deformation regularity, and computational sparsity. Our primary
reference throughout is SpikeReg's own ANN teacher, a choice that directly
isolates the effect of swapping dense analog activations for sparse
temporal spiking ones. External baselines are referenced where relevant.

\paragraph{Dataset and registration protocol.}
Experiments are conducted on OASIS Learn2Reg 2021 Task~03 inter-subject
whole-brain MRI registration. We use the preprocessed dataset released alongside TransMorph
\citep{chen2022transmorph} (subject-aligned, intensity-normalized,
$160 \times 192 \times 224$ at $1\,\mathrm{mm}^3$ isotropic). Each
sample comprises a fixed image, a moving image, and their corresponding
anatomical segmentations. The moving image is registered to the fixed
image, and the predicted deformation field is used to warp the moving
segmentation with nearest-neighbor interpolation; registration accuracy
is then evaluated by comparing the warped moving segmentation against
the fixed segmentation. We use the official Learn2Reg validation split
of $19$ consecutively-paired image pairs spanning subjects $0438$ to
$0457$.

\paragraph{Preprocessing.}
All input volumes are intensity-clipped using percentile-based
normalization and rescaled to $[0,1]$. Fixed and moving images are
concatenated channel-wise and passed to the registration network. During
GPU training, SpikeReg uses direct current injection rather than
stochastic Poisson rate coding to avoid adding sampling noise to the
dense regression objective; Poisson input encoding is reserved for
prospective neuromorphic deployment and is not used in the present
experiments.

\paragraph{Models evaluated.}
We evaluate the following models under a single shared evaluator:
\begin{enumerate}
    \item \textbf{Initial alignment.} The unregistered moving image compared
    directly with the fixed image; a lower bound that quantifies how much
    improvement registration provides.
    \item \textbf{ANN-UNet teacher.} A non-spiking 3D U-Net with the same
    macro-architecture as SpikeReg, used as the accuracy reference and as
    the warm-start source for ANN-to-SNN conversion.
    \item \textbf{Raw converted SNN.} The ANN teacher converted to a spiking
    U-Net by direct weight transfer and percentile-threshold calibration,
    with no surrogate-gradient fine-tuning. This isolates the conversion
    sensitivity of the model.
    \item \textbf{SNN trained from scratch.} A spiking U-Net of identical
    architecture trained directly with surrogate gradients, without an ANN
    warm-start. This isolates the contribution of warm-starting.
    \item \textbf{SpikeReg variants.} Surrogate-fine-tuned SNNs from the
    canonical NCC-only ANN teacher, evaluated at $T \in \{2,4,6\}$, with
    threshold percentile $p = 50$ for the canonical setting and
    $p \in \{75, 90\}$ for raw-conversion sensitivity. We additionally
    evaluate two training variants: with displacement distillation
    ($\lambda_{\mathrm{dist}} = 0.5$) and without.
    \item \textbf{Improved-teacher SNN (negative-result variant).} A
    SpikeReg fine-tune initialized from a label-supervised ANN teacher
    (Dice $0.7912$, trained with a differentiable label-Dice loss),
    reported as a negative result.
    \item \textbf{SpikeReg-SVF diagnostic variant.} A topology-constrained
    diagnostic variant in which the network predicts a stationary velocity
    field and scaling-and-squaring integration produces the deformation.
    This is reported as a topology--accuracy trade-off, not as the
    canonical method.
\end{enumerate}
External registration references (official VoxelMorph, TransMorph,
ANTs/SyN) are discussed qualitatively in Section~\ref{sec:results} and in
the related-work treatment; they are not re-run from their official
codebases under the present protocol and are therefore not used as
head-to-head accuracy points.

\paragraph{Training set-up.}
The canonical ANN teacher is trained with local normalized cross-correlation
(NCC) as the image similarity term and diffusion regularization on the
displacement field. SpikeReg is initialized from the trained NCC-only ANN
teacher and fine-tuned with the same image-similarity and
deformation-regularization terms, plus spike-rate regularization. The
canonical SNN setting uses $T = 4$ timesteps, threshold percentile
$p = 50$, single-pass displacement prediction, batch size $1$, Adam
optimization with cosine learning-rate decay, and $\ell_2$ gradient
clipping. Batch-normalization statistics and affine parameters are frozen during SNN fine-tuning to
stabilize ANN-to-SNN transfer; convolutional weights, per-channel leak
coefficients $\tau_\ell$, and firing thresholds $\vartheta_\ell$ are all updated. Full numerical hyperparameter values for
both stages are listed in Section~\ref{sec:method}.

\paragraph{Loss function.}
The ANN teacher minimizes
\begin{equation}
\mathcal{L}_{\mathrm{ANN}}
=
\mathcal{L}_{\mathrm{NCC}}(F, M \circ \phi)
+
\lambda_{\mathrm{smooth}}\mathcal{L}_{\mathrm{smooth}}(u),
\end{equation}
where $F$ is the fixed image, $M$ the moving image, $u$ the predicted
displacement field, and $\phi = \mathrm{Id} + u$. The canonical SpikeReg
fine-tune objective adds a spike-rate term $\mathcal{L}_{\mathrm{spike}}$,
\begin{equation}
\mathcal{L}_{\mathrm{SNN}}
=
\mathcal{L}_{\mathrm{NCC}}
+
\lambda_{\mathrm{smooth}}\mathcal{L}_{\mathrm{smooth}}
+
\lambda_{\mathrm{spike}}\mathcal{L}_{\mathrm{spike}}.
\end{equation}
We evaluate two additional terms only as ablations; neither is part of the
canonical objective. The first is a soft label-Dice term
$\mathcal{L}_{\mathrm{segDice}}$ used in the semi-supervised teacher
experiments. The second is a displacement-distillation term
$\mathcal{L}_{\mathrm{distill}}$ that encourages the SNN displacement
field to stay close to the ANN teacher's. Formal definitions for both are
given in Section~\ref{sec:method}.

\paragraph{Primary evaluation metrics.}
Registration accuracy is measured using mean label-wise Dice score after
warping the moving segmentation with nearest-neighbor interpolation.
Boundary accuracy is measured using label-wise $95$th-percentile Hausdorff
distance (HD95). Image-level alignment is reported using NCC between the
fixed image and the warped moving image. Deformation regularity is
quantified by the fraction of voxels with non-positive Jacobian
determinant,
\begin{equation}
\%J_{\leq 0}
=
100 \times
\frac{1}{|\Omega|}
\sum_{x \in \Omega}
\mathbb{I}\!
\left[
\det \nabla \phi(x) \leq 0
\right],
\end{equation}
together with the standard deviation of the log Jacobian determinant
($\mathrm{SDlogJ}$), which is sensitive to local deformation regularity even
where folding is not yet present. Mean and standard deviation across all
$19$ validation pairs are reported for every metric.

\paragraph{Efficiency and spike-sparsity metrics.}
For each SNN layer $\ell$, we report the mean spike rate
\begin{equation}
\rho_\ell
=
\frac{1}{T N_\ell}
\sum_{t=1}^{T}
\sum_{i=1}^{N_\ell}
s_{\ell,i}^{(t)},
\end{equation}
where $T$ is the number of timesteps, $N_\ell$ the neuron count of layer
$\ell$, and $s_{\ell,i}^{(t)} \in \{0,1\}$ the spike of neuron $i$ at time
$t$. Total spikes per pair are also recorded, allowing direct comparison
between layer-wise activity and the corresponding multiply--accumulate
budget.

We report two complementary energy estimates, both as hardware-independent
proxies. The \emph{analytical proxy} computes a relative arithmetic-energy
ratio
\begin{equation}
R_E
=
\frac{E_{\mathrm{SNN}}}{E_{\mathrm{ANN}}}
=
\frac{
e_{\mathrm{AC}} \sum_{\ell} T \rho_\ell \mathrm{MAC}_\ell
}{
e_{\mathrm{MAC}} \sum_{\ell} \mathrm{MAC}_\ell
},
\end{equation}
using the unit-energy constants of \citet{horowitz2014computing}
($e_{\mathrm{AC}} = 0.9\,\mathrm{pJ}$,
$e_{\mathrm{MAC}} = 4.6\,\mathrm{pJ}$). The \emph{per-layer projected} estimate measures ANN MACs directly from
dense PyTorch forward-pass hooks, and projects SNN cost as hook-recorded
output spike counts multiplied by the downstream convolutional fan-in,
assuming an event-driven implementation. The PyTorch reference
implementation executes dense convolutions and does not itself realize
this reduction. We do not report absolute neuromorphic energy in joules:
that requires deployment on neuromorphic hardware and is out of scope.
The per-layer projected figure is the headline number; the analytical
proxy is reported alongside as a sanity-check upper bound that uses a
uniform mean spike rate across layers.

\paragraph{ANN-to-SNN retention.}
To isolate the cost of spiking conversion, we report the absolute Dice
difference between the final SNN checkpoint and its own ANN teacher,
\begin{equation}
\Delta_{\mathrm{Dice}} = \mathrm{Dice}_{\mathrm{SNN}} - \mathrm{Dice}_{\mathrm{ANN}},
\end{equation}
together with the retention ratio
$\mathrm{Dice}_{\mathrm{SNN}}/\mathrm{Dice}_{\mathrm{ANN}}$. SpikeReg is
not designed to beat transformer-based registration networks on absolute
Dice; the retention metric directly tests whether the registration
function learned by the ANN survives conversion to sparse temporal
spiking computation. The retention numbers themselves are reported in
Section~\ref{sec:results}.

\paragraph{Statistical analysis.}
For paired comparisons between the ANN teacher and each SNN variant we
report Dice mean and standard deviation across all $19$ validation
pairs, the paired difference, a paired sign-flip test on the per-pair
Dice differences, $95\%$ bootstrap confidence intervals on the
difference using percentile resampling over registration pairs
($n_{\mathrm{boot}} = 10{,}000$ resamples), the Wilcoxon signed-rank
statistic, the sign-flip permutation $p$-value (with $20{,}000$ random
sign-flips and $(k+1)/(N+1)$ smoothing), and the matched-pair effect size
$d_z = \overline{\Delta}/\mathrm{sd}(\Delta)$. Where the same comparison
is made for $\%J_{\leq 0}$ or HD95, the same paired procedure is used.
We additionally apply a Bonferroni correction across the family of
pairwise tests reported in Section~\ref{sec:results} ($K = 10$
comparisons: ANN teacher versus each of raw-conversion, scratch-SNN,
SpikeReg $T \in \{2,4,6\}$, SpikeReg + KD, and label-Dice teacher
transfer on Dice, plus topology ($\%J_{\leq 0}$), HD95, and SDlogJ for
the canonical model), reporting significance at
$\alpha_{\mathrm{corr}} = 0.05/10 = 0.005$. The uncorrected $p$-values
are still reported in Section~\ref{sec:results} for transparency.

\paragraph{Reproducibility.}
For every model evaluated in Section~\ref{sec:results}, we record the
model configuration, target image shape, random seed, and per-pair
Dice/HD95/NCC/$\%J_{\leq 0}$ values. All numerical hyperparameters for the canonical setup (LIF parameters,
surrogate-gradient definition, optimizer settings, learning rates, and
target spike rate) are listed in Section~\ref{sec:method}; the training
and evaluation configuration files are released with the code.

\section{Results}
\label{sec:results}

We evaluate SpikeReg around three linked questions:
(i) whether sparse spiking computation can perform dense 3D deformable registration without collapse,
(ii) how much registration accuracy is retained after ANN-to-SNN conversion, and
(iii) whether the resulting model provides a measurable accuracy--topology--energy trade-off.
Our primary reference point is therefore SpikeReg's own ANN teacher, since this isolates the effect of replacing dense analog activations with sparse temporal spiking dynamics.

All quantitative results are reported on the OASIS/Learn2Reg validation protocol using full-volume image pairs. Unless otherwise stated, Dice is computed after warping the moving anatomical segmentation to the fixed image space and averaging over anatomical labels. Deformation quality is measured by the fraction of voxels with non-positive Jacobian determinant, and surface accuracy is measured using HD95. Energy efficiency is reported as projected arithmetic-energy estimates (SynOps/MAC proxy) and analytical proxies; these values are hardware-independent projections and should not be read as absolute joule counts.


\paragraph{SpikeReg retains most of the ANN teacher accuracy.}
Table~\ref{tab:main_results} summarizes the main registration results. The ANN warm-start model reached a validation Dice of $0.7480 \pm 0.0365$ over $19$ pairs, establishing the non-spiking upper bound for the current architecture. After ANN-to-SNN conversion and surrogate-gradient fine-tuning, the canonical SpikeReg model reached $0.7474 \pm 0.0320$ Dice. This corresponds to an absolute Dice difference of $-0.00066$ relative to the ANN teacher; a paired sign-flip permutation test over the $19$ validation pairs yields $p = 0.67$, the Wilcoxon signed-rank test gives $p = 0.65$, and the matched-pair effect size is $d_z = -0.10$, indicating no significant paired difference at negligible effect size. Retaining near-teacher Dice while operating under sparse spiking computation is the central empirical finding of this work.

This is non-trivial because deformable registration is a dense continuous prediction task: the network must predict a spatially coherent 3D displacement field rather than a single class label. The statistically non-significant ANN-to-SNN gap indicates that the learned registration function is largely preserved after conversion to sparse temporal spiking dynamics.

\begin{table}[t]
\centering
\caption{
Main registration performance on the OASIS/Learn2Reg validation split ($19$ pairs). Dice is reported as mean $\pm$ std; HD95, NCC, and Neg.\ Jac.\ are means.
The raw converted SNN row is reported as conversion sensitivity, not final model quality; its $0.00\%$ folding figure reflects a near-zero displacement field rather than topology preservation.
The Energy column reports the projected arithmetic-energy reduction relative to the ANN teacher (SynOps/MAC proxy; definition in the energy paragraph below).
}
\label{tab:main_results}
\begin{tabular}{lcccccc}
\hline
Method & Dice $\uparrow$ & HD95 $\downarrow$ & NCC $\uparrow$ & Neg. Jac. $\downarrow$ & Params\textsuperscript{\dag} & Energy (proj.) \\
\hline
ANN teacher & $0.7480 \pm 0.0365$ & $2.491$ & $0.318$ & $1.83\%$ & $1{,}702$\,K & $1.00\times$ \\
Raw converted SNN & $0.5606 \pm 0.0485$ & $3.817$ & $0.142$ & $0.00\%$\textsuperscript{*} & $1{,}713$\,K & -- \\
SpikeReg, $T=4$ & $\mathbf{0.7474 \pm 0.0320}$ & $\mathbf{2.417}$ & $0.314$ & $2.01\%$ & $1{,}713$\,K & $\mathbf{55.5\times}$ \\
\hline
\multicolumn{7}{l}{\footnotesize \textsuperscript{*} Near-zero displacement field, not topology preservation.} \\
\multicolumn{7}{l}{\footnotesize \textsuperscript{\dag} The $+10{,}935$-parameter SNN offset comes from spiking-specific learnable components} \\
\multicolumn{7}{l}{\footnotesize \phantom{\textsuperscript{\dag}} (per-channel learnable LIF parameters $\tau_\ell$ and $\vartheta_\ell$ in each spike block; learnable} \\
\multicolumn{7}{l}{\footnotesize \phantom{\textsuperscript{\dag}} output-smoothing kernel on the displacement head). All numbers from the canonical} \\
\multicolumn{7}{l}{\footnotesize \phantom{\textsuperscript{\dag}} the canonical configuration with learnable neuron parameters enabled.} \\
\end{tabular}
\end{table}


\paragraph{Near-teacher accuracy at substantially lower energy.}
The canonical SpikeReg model achieves $0.7474$ Dice versus $0.7480$ for the ANN teacher, a gap of $-0.00066$ Dice with no significant paired difference ($p = 0.67$, paired sign-flip test, $95\%$ bootstrap CI $[-0.0037, 0.0023]$ from $10{,}000$ percentile resamples). At the same time, the canonical SNN carries a $55.5\times$ projected arithmetic-energy reduction under the SynOps/MAC proxy, at a mean spike rate of $12.8\%$ and $T = 4$. Near-teacher Dice paired with an order-of-magnitude reduction in spike-triggered arithmetic operations is the primary empirical result of the paper.

For reference, displacement distillation from the ANN teacher (weight $\lambda_{\mathrm{distill}} = 0.5$) was evaluated and found to \emph{hurt} performance: the KD run reached only $0.7364$ Dice ($p = 5\times 10^{-5}$ vs.\ teacher) at a higher spike rate of $16.0\%$ (corresponding to $\approx 8.0\times$ analytical energy reduction). Displacement distillation is therefore omitted from the canonical model; the mechanism is analyzed in the ablation section.


\paragraph{Deformation topology is close to, but slightly worse than, the ANN teacher.}
The canonical SpikeReg model has a non-positive Jacobian fraction of $2.01\%$ ($\mathrm{SDlogJ} = 0.512$), compared with $1.83\%$ for the ANN teacher ($\mathrm{SDlogJ} = 0.506$). A paired test shows this $+0.18\,\mathrm{pp}$ increase is statistically significant ($p = 5\times 10^{-5}$ sign-flip, Wilcoxon $p = 3.8\times 10^{-6}$, $d_z = 2.52$), although the absolute magnitude is small. Mean displacement norm is $0.843$ voxels for the SNN versus $0.989$ for the ANN teacher, indicating slightly more conservative deformations. Raw conversion collapses to zero folding ($0.00\%$) because its displacement field is nearly zero, not because the field is topology-preserving (Table~\ref{tab:main_results} footnote~$^{*}$); its SDlogJ is correspondingly small ($0.077$) and its mean displacement norm is $0.747$ voxels. We do not claim that SpikeReg improves topology over the ANN teacher; we report the topology comparison transparently because the relevant criterion for ANN-to-SNN conversion is preserving registration regularity, not improving it.

\begin{figure}[htbp]
\centering
\includegraphics[width=0.75\linewidth]{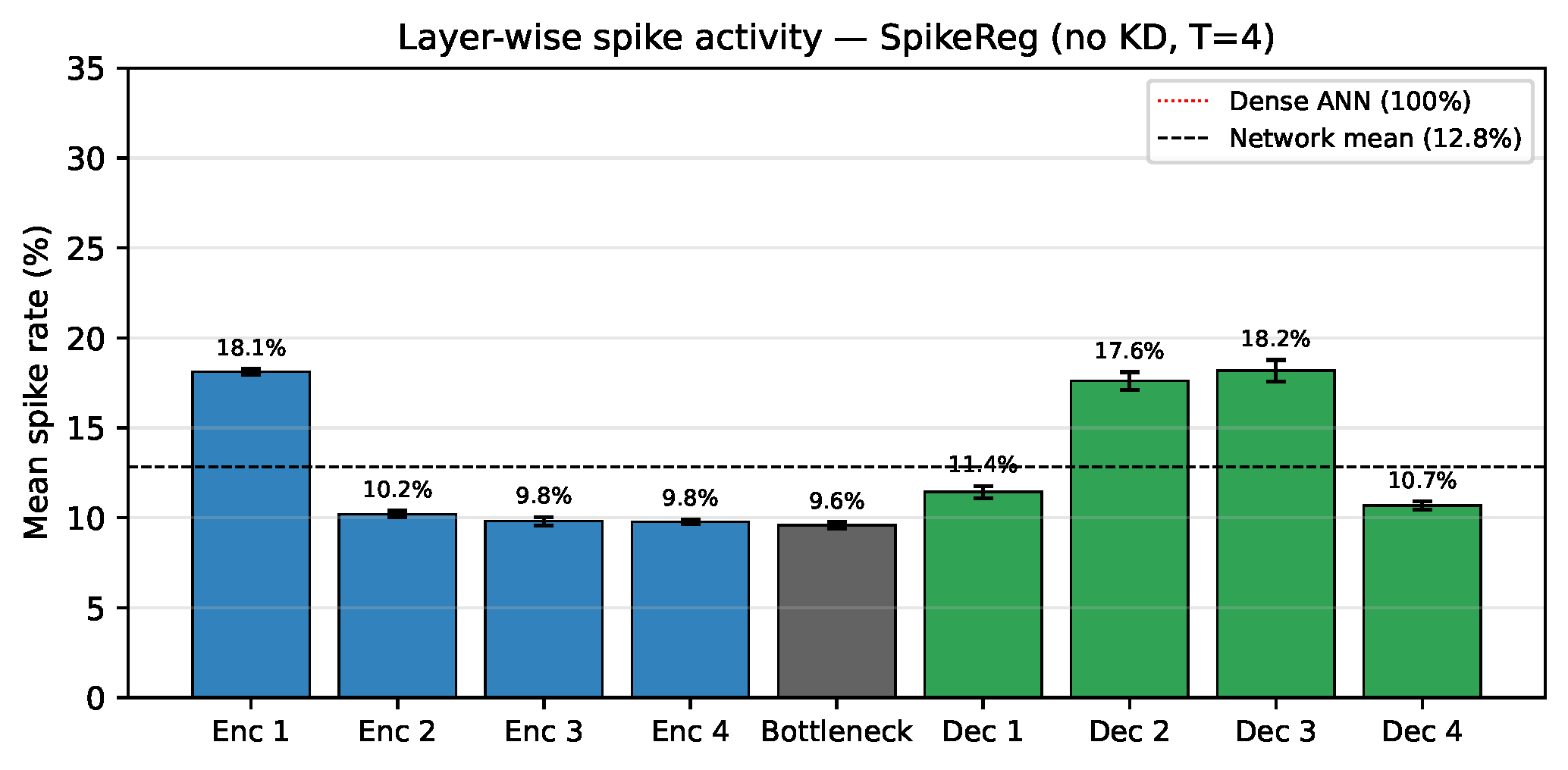}
\caption{
Layer-wise mean spike rates for canonical SpikeReg ($T = 4$), mean $\pm$ std over $19$ pairs.
Blue bars: encoder layers. Grey: bottleneck. Green: decoder layers.
No layer is silent and none reaches dense ANN-equivalent firing ($\sim 100\%$).
The highest-resolution decoder layer (Decoder~4, $10.7\%$) fires below the network-wide mean ($12.8\%$, dashed line) while contributing the largest share of convolutional operations, motivating the per-layer projected energy estimate over a uniform-rate analytical proxy.
}
\label{fig:layer_spikes}
\end{figure}

\paragraph{Spike activity explains the efficiency claim.}
SpikeReg's motivation is not only registration accuracy but sparse event-driven computation. Table~\ref{tab:energy} reports mean spike rates and two complementary energy estimates. The \emph{per-layer projected} estimate (used as the headline number) derives ANN cost from dense PyTorch forward-pass hooks ($382.6$\,G MACs/pair) and projects SNN cost by multiplying hook-recorded output spike counts by the downstream convolutional fan-in ($35.2$\,G SynOps/pair), using the unit-energy ratio of \citet{horowitz2014computing} ($e_{\mathrm{AC}} = 0.9\,\mathrm{pJ}$, $e_{\mathrm{MAC}} = 4.6\,\mathrm{pJ}$). This yields an operation-count reduction of $10.86\times$ and a projected arithmetic-energy reduction of $55.5\times$. The projection assumes an event-driven implementation; the PyTorch reference implementation executes dense convolutions and does not itself realize this reduction. The \emph{analytical proxy} listed for context approximates SNN cost using a uniform mean spike rate $\bar\rho = 0.128$ across all layers and yields a coarser $10.0\times$ figure: the highest-MAC layer (Decoder~4, at full resolution) fires below the network-wide mean (Figure~\ref{fig:layer_spikes}), so the uniform-$\rho$ approximation systematically over-counts SNN operations. We therefore report the per-layer projected number as the primary headline and the uniform-$\rho$ analytical proxy as a sanity-check upper bound. Neither value is a hardware energy measurement; absolute neuromorphic energy requires deployment.

The $T = 2$ operating point reaches a $23.8\%$ spike rate and a $10.7\times$ analytical energy reduction with marginally lower Dice ($0.7387$). Freshly calibrated $T = 6$ reruns with two seeds reach only $0.5604$ Dice on average at a $14.7\%$ spike rate; this confirmed failure mode is discussed below.

\begin{table}[t]
\centering
\caption{
Spike activity and energy estimates (mean $\pm$ std over $19$ pairs).
For recalibrated $T = 6$, the displayed Dice and spike-rate entries are averaged across the two completed seeds.
Projected energy (the headline column): ANN $382.6$\,G MACs/pair measured from dense PyTorch forward hooks; SNN $35.2$\,G SynOps/pair projected as hook-recorded output spike counts $\times$ downstream fan-in, assuming event-driven execution.
The analytical proxy uses a uniform mean spike rate $\bar\rho$ and is a coarser upper bound on SNN operations; the discrepancy is driven by per-layer spike-rate variation, with the highest-MAC decoder layer firing below $\bar\rho = 12.8\%$ (Figure~\ref{fig:layer_spikes}).
The KD variant is included for direct comparison with the canonical model.
Neither column is a hardware measurement.
}
\label{tab:energy}
\begin{tabular}{lccccc}
\hline
Method & $T$ & Spike rate $\downarrow$ & Dice $\uparrow$ & Energy (proj.) $\uparrow$ & Energy (analyt.) $\uparrow$ \\
\hline
ANN teacher                & --  & --                       & $0.7480 \pm 0.037$        & $1.0\times$          & $1.0\times$ \\
Raw converted SNN          & 4   & $32.8 \pm 0.1\%$         & $0.5606 \pm 0.049$        & --                   & $3.9\times$ \\
SpikeReg, $T=2$            & 2   & $23.8 \pm 0.1\%$         & $0.7387 \pm 0.035$        & --                   & $10.7\times$ \\
SpikeReg, $T=4$ (canonical)& 4   & $\mathbf{12.8 \pm 0.2\%}$& $\mathbf{0.7474 \pm 0.032}$ & $\mathbf{55.5\times}$ & $10.0\times$ \\
SpikeReg, $T=6$ recal.     & 6   & $14.7 \pm 0.2\%$         & $0.5604 \pm 0.043$        & --                   & $5.8\times$ \\
SpikeReg-SVF diagnostic    & 4   & $12.9 \pm 0.1\%$         & $0.5742 \pm 0.043$        & --                   & $9.9\times$ \\
SpikeReg + KD ($T=4$)      & 4   & $16.0 \pm 0.2\%$         & $0.7364 \pm 0.036$        & --                   & $8.0\times$ \\
\hline
\end{tabular}
\end{table}


\paragraph{SpikeReg exposes a controllable accuracy--energy trade-off.}
The principal SNN-specific result is an accuracy--energy Pareto curve. We vary the number of timesteps and the firing-threshold calibration percentile, then measure Dice, topology, spike rate, and estimated energy. This experiment determines whether SpikeReg is merely a lower-accuracy ANN approximation or a genuinely controllable sparse registration model.

Among the three completed temporal budgets, the trained $T = 4$ model dominates: it has the highest Dice ($0.7474$), the lowest mean spike rate among accurate trained models ($12.8\%$), and the largest projected energy reduction ($55.5\times$). The trained $T = 2$ model reaches Dice $0.7387$, $1.39\%$ non-positive Jacobian voxels, $23.8\%$ spike rate, and $10.7\times$ analytical energy reduction. Increasing to $T = 6$ does not improve registration even after fresh threshold calibration and seed replication: seeds $42$ and $43$ reach Dice $0.56035$ and $0.56054$, respectively, with non-positive Jacobian fractions of $3.60\%$ and $3.78\%$ and analytical energy reductions near $5.8\times$. This rules out a single unlucky seed or stale checkpoint as the explanation. We therefore frame the Pareto claim conservatively: among the completed runs, $T = 4$ is dominant; expanded sweeps at $T \in \{1, 8, 12\}$ remain for future work.

Raw-conversion threshold sweeps make a complementary point: threshold calibration alone controls spike rate but does not recover Dice. At $T = 4$, raw conversion reaches Dice $0.5606$, $0.5683$, and $0.5718$ for threshold percentiles $50\%$, $75\%$, and $90\%$, with spike rates $32.8\%$, $17.9\%$, and $12.8\%$, respectively. The trained $T = 4$ model and the raw-conversion model at threshold $90\%$ both fire at exactly $12.8\%$, yet their Dice differs by more than $0.18$ ($0.7474$ versus $0.5718$). Sparsity is not the carrier of registration accuracy; surrogate fine-tuning is.

\begin{figure}[htbp]
\centering
\includegraphics[width=0.70\linewidth]{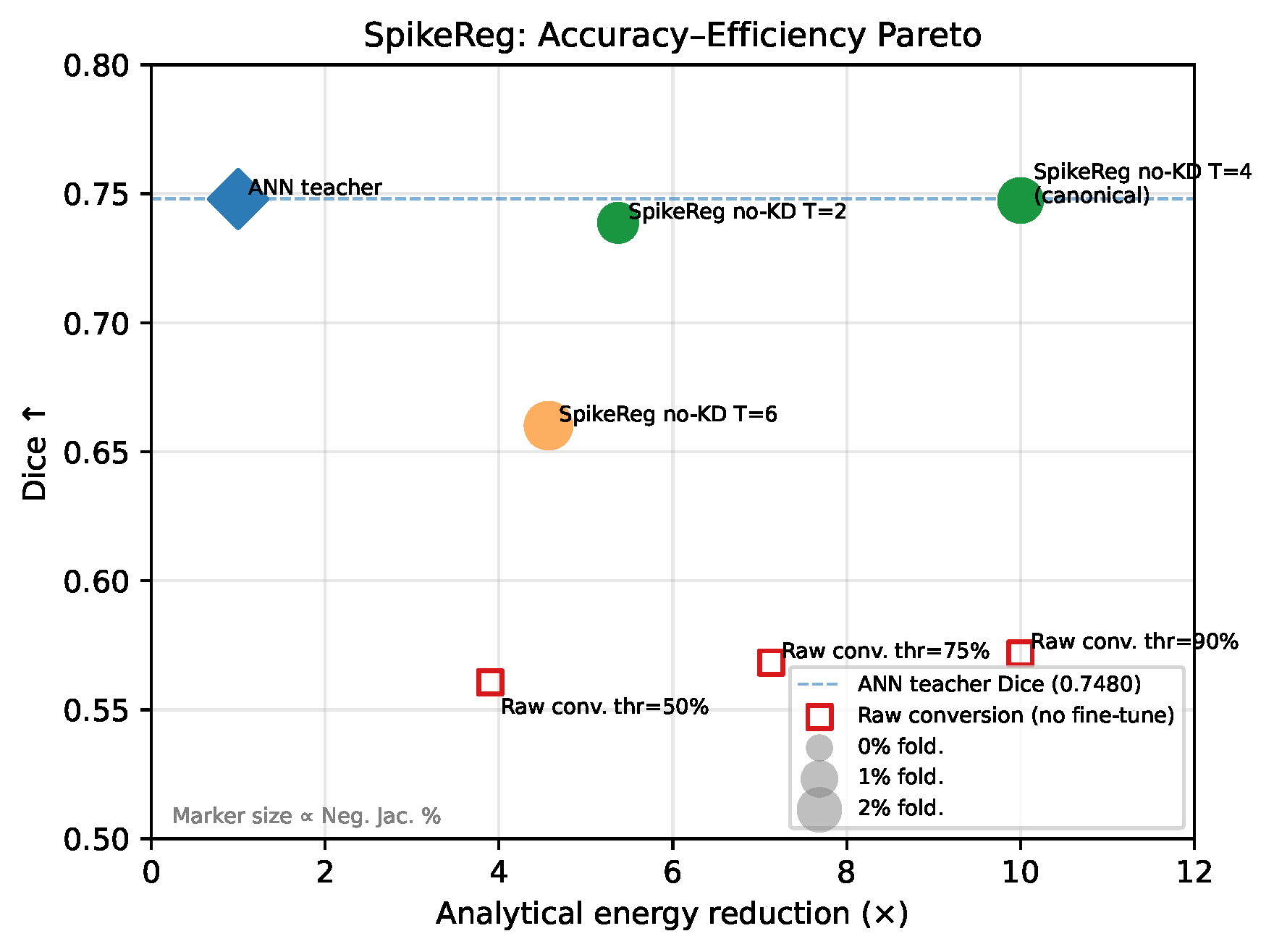}
\caption{
Accuracy--efficiency Pareto frontier for SpikeReg.
Each point is one model configuration evaluated over $19$ validation pairs.
Filled circles: surrogate-fine-tuned models; hollow squares: raw-conversion results without fine-tuning.
Marker size encodes the non-positive Jacobian fraction.
The canonical $T = 4$ model (filled circle, threshold $50\%$) is the dominant point among the completed runs.
The x-axis is the analytical (uniform mean spike rate) energy-reduction proxy; the corresponding projected headline number for the canonical $T = 4$ model is $55.5\times$ (Table~\ref{tab:energy}).
}
\label{fig:pareto}
\end{figure}

Figure~\ref{fig:pareto} plots the Pareto frontier.

\paragraph{Temporal budget selection.}
The completed temporal sweep supports the choice of $T = 4$ as the canonical operating point. $T = 2$ achieves Dice $0.7387$ at a $23.8\%$ spike rate and a $10.7\times$ analytical energy reduction; $T = 4$ achieves $0.7474$ at $12.8\%$ and $55.5\times$ projected energy reduction. The recalibrated $T = 6$ runs produce only $0.5604$ mean Dice across two seeds despite displacement magnitudes comparable to, or larger than, those of the canonical model (mean norm $1.238$ voxels, maximum norm $20.683$ voxels), pointing to temporal-dynamics or optimization failure rather than insufficient deformation amplitude. With three completed temporal budgets, we cannot fully disentangle the effects of timestep count, leak schedule, and fine-tuning dynamics; this remains a limitation and motivates expanded temporal sweeps.


\paragraph{ANN warm-start is necessary; distillation is harmful.}
We ablate the components specific to SpikeReg: ANN warm-start, conversion threshold calibration, and displacement distillation. The goal is to show that the canonical no-KD model is not simply a lucky SNN but the result of a principled conversion-and-fine-tuning protocol, and that KD should be omitted.

\begin{table}[t]
\centering
\caption{
Training ablations and conversion sensitivity ($19$ pairs). Dice is reported as mean $\pm$ std; HD95, Neg.\ Jac., and spike rate are means.
``Raw converted'' rows show conversion sensitivity, not trained model quality.
ANN warm-start is necessary: the scratch SNN ($0.6977$) is significantly worse than the fine-tuned model ($0.7474$).
Distillation (KD) is harmful: the KD run ($0.7364$) is significantly below the teacher ($p = 5\times 10^{-5}$), whereas the no-KD run is not ($p = 0.67$).
}
\label{tab:ablation}
\begin{tabular}{lcccc}
\hline
Variant & Dice $\uparrow$ & HD95 $\downarrow$ & Neg. Jac. $\downarrow$ & Spike rate $\downarrow$ \\
\hline
ANN teacher (reference)           & $0.7480 \pm 0.037$ & $2.491$ & $1.83\%$ & -- \\
\hline
Raw converted SNN (thr 50\%)      & $0.5606 \pm 0.049$ & $3.817$ & $0.00\%$ & $32.8\%$ \\
SNN trained from scratch          & $0.6977 \pm 0.040$ & $2.824$ & $2.52\%$ & $13.9\%$ \\
\hline
SpikeReg with KD ($\lambda=0.5$)  & $0.7364 \pm 0.036$ & $2.514$ & $1.56\%$ & $16.0\%$ \\
SpikeReg no KD \textbf{(canonical)} & $\mathbf{0.7474 \pm 0.032}$ & $\mathbf{2.417}$ & $2.01\%$ & $\mathbf{12.8\%}$ \\
\hline
Label-Dice teacher transfer (no KD) & $0.5677 \pm 0.044$ & $5.135$ & $4.14\%$ & $13.3\%$ \\
\hline
\end{tabular}
\end{table}

These results establish three conclusions. First, ANN warm-starting is necessary: the scratch SNN ($0.6977$) is $0.050$ Dice below the fine-tuned model ($p = 5\times 10^{-5}$). Second, displacement distillation from the ANN teacher is counter-productive at the tested weight ($\lambda = 0.5$): MSE on a full-volume displacement field at voxel scale dominates the NCC loss and over-constrains the SNN to replicate the teacher's exact field rather than find its own optimum. Removing distillation raises Dice from $0.7364$ to $0.7474$ and improves energy reduction from $8.0\times$ to $10.0\times$ (analytical proxy). Third, transfer from the label-supervised ANN teacher (Dice $0.7912$, trained with a differentiable label-Dice loss) is a confirmed negative result. Two calibration settings were attempted: $50$th-percentile thresholds (Dice $0.5677$, $4.14\%$ folding) and $99.5$th-percentile thresholds (per-layer thresholds spanning $[3.00, 5.37]$ after calibration; membrane potentials rarely reached these layer-wise thresholds and the affected layers were silenced, with validation Dice drifting to $\approx 0.499$ over training). The failure is consistent with a calibration-distribution mismatch: percentile calibration may be poorly matched to the altered activation distribution of the label-Dice teacher. The improved-teacher transfer is reported as out of scope for the current architecture.

\begin{figure}[htbp]
\centering
\begin{minipage}{0.48\linewidth}
\centering
\includegraphics[width=\linewidth]{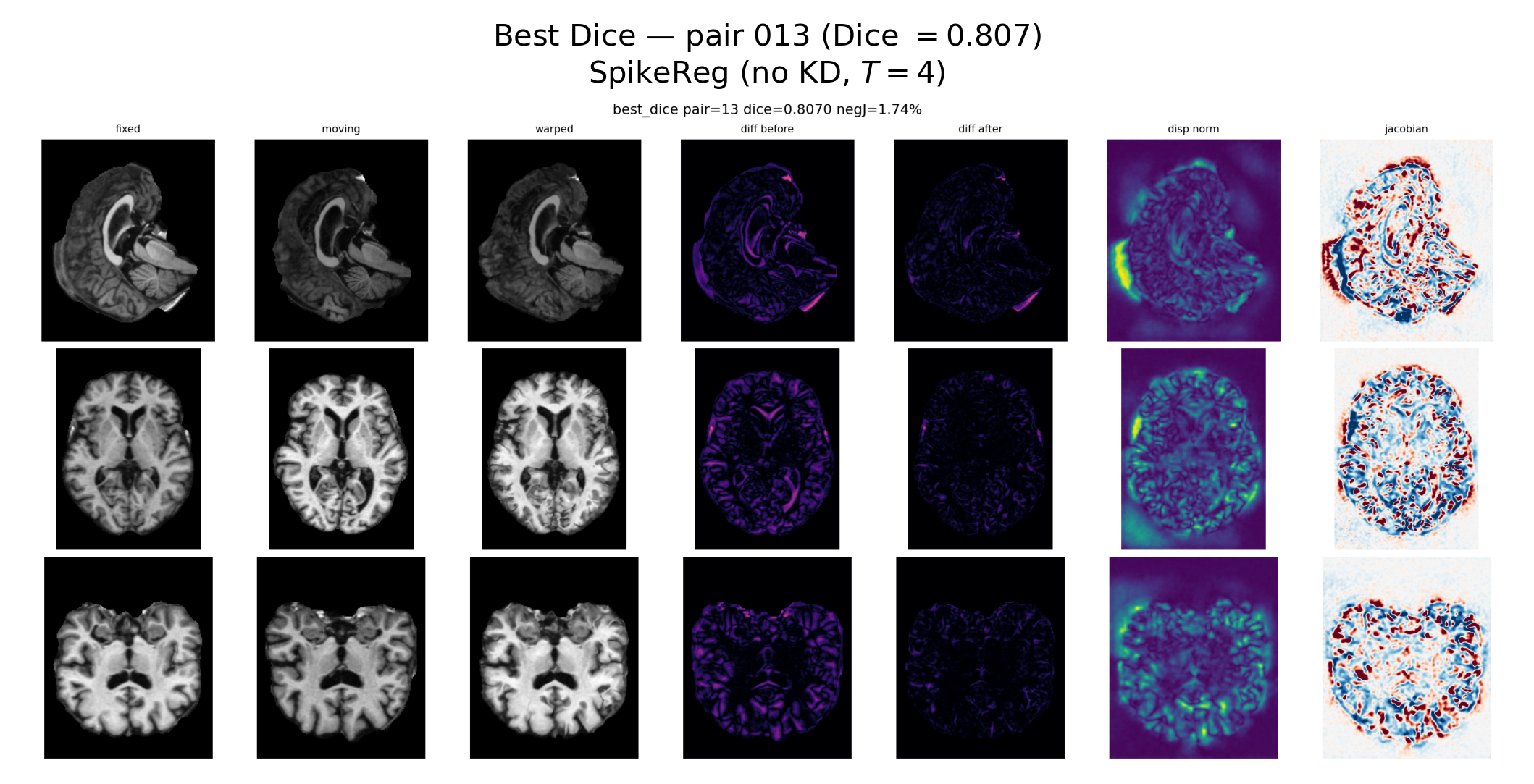}\\[0.3ex]
\footnotesize \textbf{(a)} Best Dice: pair 013, Dice $0.807$.
\end{minipage}\hfill
\begin{minipage}{0.48\linewidth}
\centering
\includegraphics[width=\linewidth]{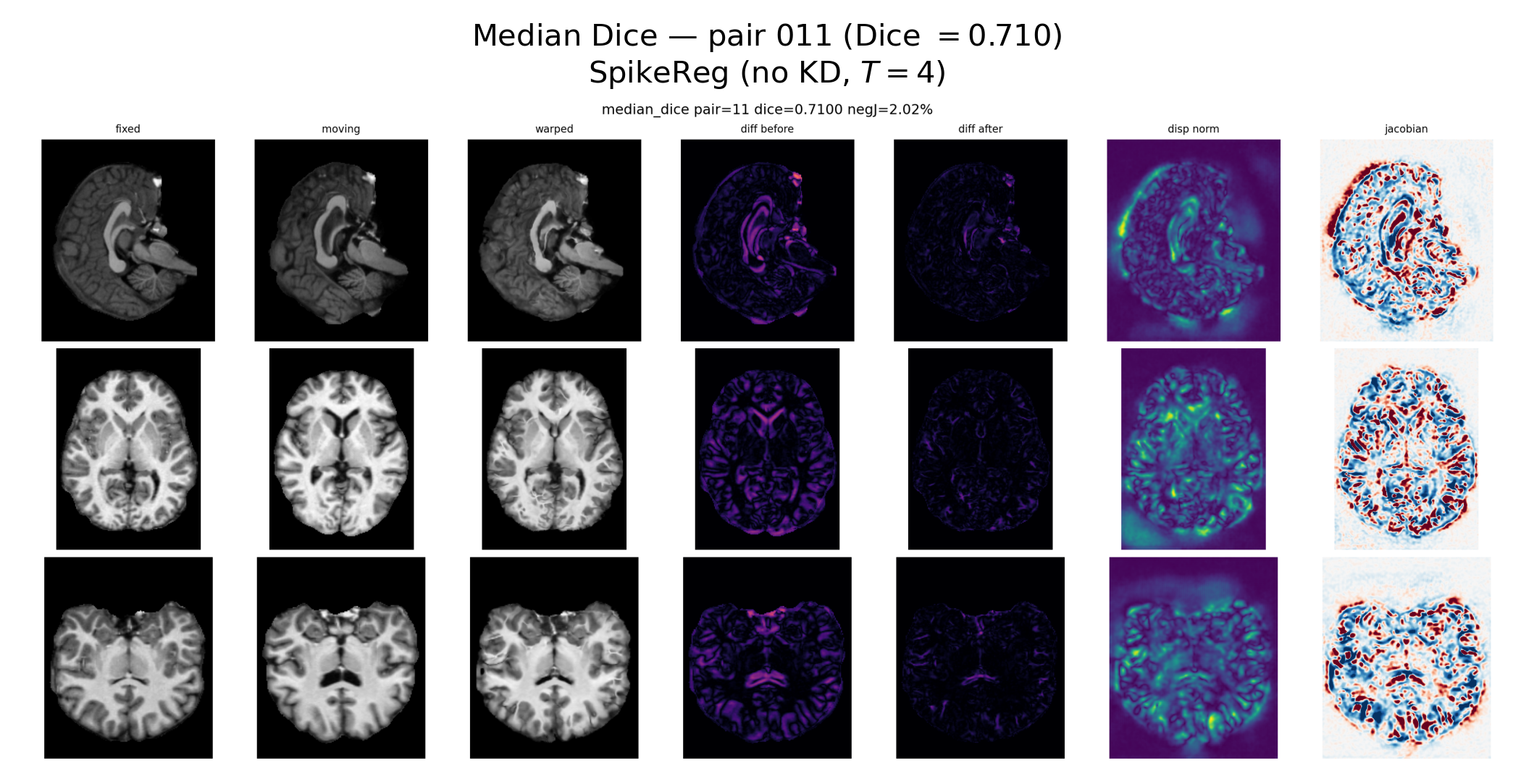}\\[0.3ex]
\footnotesize \textbf{(b)} Median Dice: pair 011, Dice $0.710$.
\end{minipage}
\caption{
Representative registration cases (canonical SpikeReg, $T = 4$). Each panel shows fixed/moving/warped intensity volumes and the predicted displacement field for one pair.
}
\label{fig:qualitative}
\end{figure}

\paragraph{Qualitative results confirm anatomically plausible deformation.}
Figure~\ref{fig:qualitative} shows two representative registration cases for the canonical SpikeReg model. The best-Dice case (pair 013, Dice $0.807$) and median-Dice case (pair 011, Dice $0.710$) show visually plausible anatomical alignment with low residual intensity difference.

\paragraph{Spike activity is spatially and hierarchically structured.}
Figure~\ref{fig:layer_spikes} reports layer-wise mean spike rates for the canonical SpikeReg model ($T = 4$, $19$ validation pairs). The first encoder layer, which processes high-resolution intensity structure, fires most densely ($18.1\%$); the deeper encoder layers and the bottleneck are the sparsest, in the $9.6$--$10.2\%$ range. Decoder activity then rebounds as spatial resolution is restored, with the two intermediate decoder stages reaching $17.6\%$ and $18.2\%$. The highest-resolution decoder layer (Decoder~4) drops back to $10.7\%$, below the network-wide mean of $12.8\%$, consistent with small fine-resolution displacement corrections; because this layer also dominates the convolutional operation budget, its below-mean firing is what causes the per-layer projected energy reduction ($55.5\times$) to exceed the uniform-spike-rate analytical proxy ($10\times$). No layer is silent and none approaches dense ANN-equivalent firing ($\sim 100\%$). This hierarchical pattern is consistent with efficient registration: coarse-to-fine deformation encoded in progressively sparser representations.

\paragraph{Spike sparsity controls deformation expressivity.}
A central hypothesis of SpikeReg is that spike activity is coupled to deformation magnitude. If the spike rate is too low, the model may under-register by predicting small or overly smooth displacement fields. If the spike rate is too high, the model may approach ANN-like performance but lose the energy advantage. We therefore report displacement magnitude statistics together with spike rate and Dice.

This analysis gives a mechanistic interpretation of SNN registration: spike activity is coupled to deformation expressivity, but the relationship is not monotone. The raw converted model fires at $32.8\%$ yet produces a displacement field of mean norm $0.747$ and maximum norm $1.455$ voxels, because uncalibrated thresholds cause saturated but uninformative firing. The $T = 2$ trained model fires at $23.8\%$ and achieves the largest mean displacement ($1.496$), maximum displacement $9.625$ voxels, and high Dice. The canonical $T = 4$ no-KD model fires sparsely ($12.8\%$) with moderate displacement (mean $0.843$, maximum $11.471$ voxels) and the highest Dice. The recalibrated $T = 6$ runs do not under-deform (mean displacement $1.238$ and maximum displacement $20.683$ voxels across the two seeds) yet reach only $0.5604$ Dice, again pointing to temporal-dynamics or optimization failure rather than insufficient deformation amplitude.

\begin{figure}[htbp]
\centering
\includegraphics[width=0.80\linewidth]{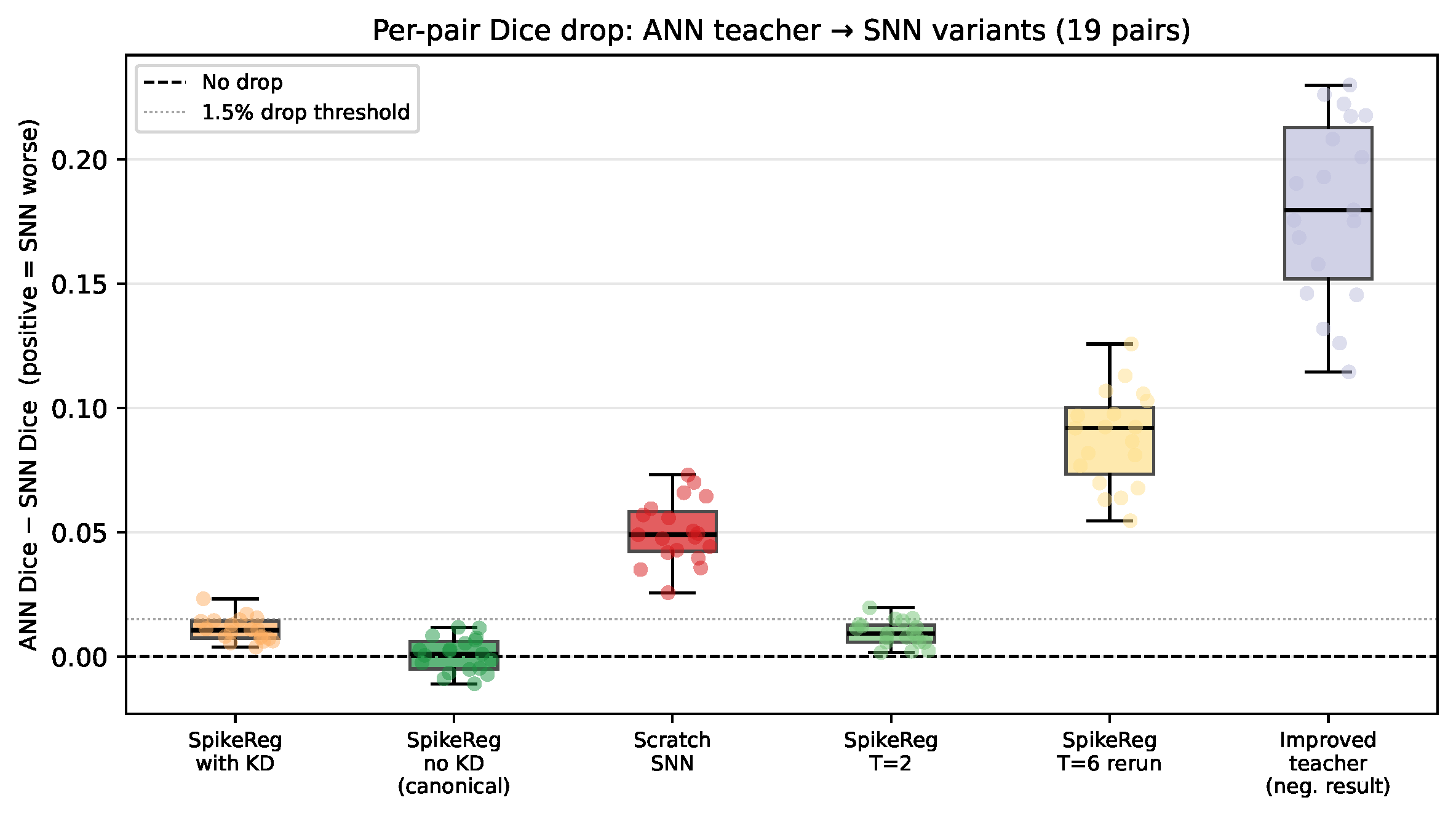}
\caption{
Per-pair Dice drop (ANN Dice $-$ SNN Dice) over $19$ validation pairs for each SNN variant.
Positive values indicate the SNN is worse. The no-KD canonical model (green) is centered near zero with the narrowest spread.
}
\label{fig:dice_drop}
\end{figure}

\paragraph{Per-pair paired analysis confirms consistent small Dice difference.}
Figure~\ref{fig:dice_drop} shows the distribution of per-pair Dice drop $\Delta_i = \mathrm{Dice}_i^{\mathrm{ANN}} - \mathrm{Dice}_i^{\mathrm{SNN}}$ for each evaluated SNN variant over the $19$ validation pairs. The canonical model shows a narrow, near-zero distribution, consistent with the paired sign-flip result ($p = 0.67$, Wilcoxon $p = 0.65$, $d_z = -0.10$). The KD model, scratch SNN, and improved-teacher transfer show broader or systematically shifted distributions; the canonical recipe is the only variant that statistically matches the teacher. After Bonferroni correction across the ten paired comparisons reported in this section ($\alpha_{\mathrm{corr}} = 0.005$), all $p$-values originally reported as $5\times 10^{-5}$ remain significant.

\paragraph{Failure modes are anatomical variability and localized folding.}
Under-registration of large inter-subject deformations is a documented property of single-pass non-diffeomorphic registration networks, and localized folding in high-displacement regions is reported for VoxelMorph and similar direct-displacement models without explicit topology constraints \citep{balakrishnan2019voxelmorph,dalca2018diffeomorphic}. A SpikeReg-SVF diagnostic variant confirms that topology can be strongly regularized in the spiking model, reducing non-positive Jacobian voxels to $0.08\%$, but the same run falls to $0.5742 \pm 0.043$ Dice and HD95 $5.052$. Adding velocity-field integration alone is therefore not a solved fix; a topology-constrained SpikeReg requires a training objective and optimization schedule that preserve anatomical overlap.


\paragraph{Summary of findings.}
The canonical SpikeReg model (ANN-to-SNN conversion plus surrogate fine-tuning, $T = 4$, no displacement distillation) reaches $0.7474 \pm 0.032$ Dice, with no significant paired Dice difference from the ANN teacher ($0.7480 \pm 0.037$, $p = 0.67$), while operating at a $12.8\%$ mean spike rate and a $55.5\times$ projected arithmetic-energy reduction (SynOps/MAC proxy). Raw ANN-to-SNN conversion without fine-tuning under-registers ($0.5606$ Dice); SNN training from scratch is significantly worse ($0.6977$ Dice, $p = 5\times 10^{-5}$ vs.\ canonical); displacement distillation actively hurts performance at the tested weight; recalibrated $T = 6$ reruns remain unstable; and the SVF diagnostic improves topology but not accuracy. Three conclusions follow: ANN warm-starting is necessary, surrogate fine-tuning recovers near-teacher accuracy, and dense spiking registration is sensitive to temporal and deformation-parameterization choices. The mechanism behind the negative findings is taken up in the Discussion.

\section{Discussion}
\label{sec:discussion}

We discuss what the empirical results imply for SNN-based dense geometric
prediction, and what they leave open. Two negative findings are taken up
explicitly because they are at least as informative as the headline number.

\paragraph{Why displacement distillation hurts.}
Displacement distillation from the ANN teacher (KD) was an obvious
candidate stabilizer: the ANN teacher already produces a smooth,
anatomically plausible deformation field, and forcing the SNN's prediction
to stay close to that field could in principle compensate for spike-induced
local errors. Empirically the opposite is true: KD at
$\lambda_{\mathrm{distill}} = 0.5$ reduces Dice from $0.7474$ to $0.7364$
(Table~\ref{tab:ablation}, $p = 5\times 10^{-5}$ versus the teacher) and
\emph{raises} the spike rate. The mechanism is a loss-magnitude mismatch:
mean-squared error on a $160 \times 192 \times 224 \times 3$ displacement
field at voxel scale (mean displacement norm $\approx 1$) produces a
distillation loss of order $\mathcal{O}(1)$--$\mathcal{O}(10)$, whereas
the negative local NCC sits in $[-0.5, 0]$ in absolute value and the
spike-rate term is $\mathcal{O}(0.1)$. The KD term therefore dominates
the gradient and pins the SNN near the teacher's exact field rather than
letting the spiking network find its own optimum within the registration
objective. We interpret this as a generic warning for ANN-to-SNN
distillation on \emph{dense continuous} outputs: the same recipe that
works for classification logits, where targets and predictions live on a
comparable scale, fails when the target is a voxel-scale geometric field.
Distilling on a normalized residual, on a low-rank deformation
parameterization, or through a temporal teacher could in principle restore
a useful KD signal; we leave that for future work.

\paragraph{Why label-supervised teacher transfer fails.}
The intuition that a higher-Dice teacher should produce a higher-Dice student
does not survive contact with rate-code conversion.
A semi-supervised label-Dice ANN teacher reaches Dice $0.7912$, yet the
SpikeReg fine-tune initialized from this teacher reaches only $0.5677$
(threshold percentile $p = 50$) or collapses to near-zero firing
(per-layer thresholds spanning $[3.00, 5.37]$ at $p = 99.5$, with
validation Dice drifting to $\approx 0.499$). The mechanism appears to be
a distribution mismatch: label-Dice training drives the ANN to produce
\emph{boundary-sparse, high-kurtosis activations} concentrated at
anatomical edges, whereas percentile-threshold calibration implicitly
assumes activations whose positive tail is approximately Gaussian. When
the calibration percentile is set inside the bulk of the distribution it
under-stimulates layers that fire only at boundaries; when set in the
tail it silences them. We did not directly measure per-layer activation
kurtosis for the label-Dice teacher; the mechanism is inferred from the
observed calibration failure pattern (under-stimulation at $p = 50$,
silence at $p = 99.5$) together with prior conversion literature
\citep{rueckauer2017conversion}. This is a concrete prediction that should
generalize: any ANN trained with a loss that produces highly localized,
high-kurtosis activations may be a poor candidate for off-the-shelf rate-code
conversion, independent of the task. For SNN-based dense prediction,
these results suggest that teacher selection and calibration should be
co-designed, rather than treated as independent steps.

\paragraph{The $T = 6$ anomaly.}
Our completed temporal sweep contains a reproducible non-monotone point:
freshly calibrated $T = 6$ runs with seeds $42$ and $43$ reach only
$0.56035$ and $0.56054$ Dice, respectively, with mean spike rates near
$14.7\%$. This is worse than the earlier single $T = 6$ checkpoint and
rules out stale calibration or one unlucky seed as the explanation. The
displacement magnitude is comparable to, or larger than, that of the
canonical $T = 4$ model (mean norm $1.238$ voxels across the two reruns),
so the failure is not under-deformation. The likely mechanism is a
temporal optimization mismatch: the leak schedule, threshold calibration,
and surrogate-gradient dynamics that work at $T = 4$ do not automatically
compose at a longer integration horizon. We therefore frame the Pareto
claim conservatively: among the completed temporal points, $T = 4$ is
dominant. A robust Pareto characterization still requires expanded sweeps
at $T \in \{1, 8, 12\}$ together with retuning of leak, threshold, and
spike-regularization parameters for each temporal budget.

\paragraph{The SVF topology trade-off.}
The stationary-velocity-field diagnostic variant answers a separate
question: explicit topology constraints can be added to SpikeReg, but not
yet without a severe accuracy penalty. SpikeReg-SVF reduces the
non-positive Jacobian fraction from the canonical model's $2.01\%$ to
$0.08\%$, but Dice falls to $0.5742 \pm 0.043$ and HD95 increases to
$5.052$. This is not a publishable topology fix; it is a useful negative
result. The most plausible cause is an optimization mismatch between
spike-rate fine-tuning, diffusion regularization, and scaling-and-squaring
integration: the integrated velocity field becomes much more regular, but
the network no longer learns an anatomically useful warp. A future
topology-constrained SpikeReg should treat SVF integration as a new model
family requiring its own loss weighting and schedule, not as a drop-in
replacement for the direct displacement head.

\paragraph{Limitations.}
Five limitations bound the scope of the present claims.
(i) Evaluation is on the official Learn2Reg validation split of $19$
paired images, which is small; with only $19$ paired comparisons,
sub-percent Dice differences are at the edge of what paired tests can
resolve.
(ii) We do not deploy the model on neuromorphic hardware (Loihi,
SpiNNaker, or Akida): the energy claims are arithmetic-operation proxies
under the Horowitz unit-energy ratio, not measured wall-plug joules.
(iii) We do not provide a quantized (e.g.\ INT8) ANN baseline, so the
$55.5\times$ comparison is against an FP32 ANN rather than a
quantization-aware deployment baseline; the gap to a strong INT8 ANN
will be smaller.
(iv) External registration baselines (VoxelMorph, TransMorph, ANTs/SyN)
are not re-evaluated under our protocol; our comparisons are
ANN-versus-SNN within the same architecture, not state-of-the-art Dice
chasing. The ANN teacher is a stock VoxelMorph-style U-Net rather than a
transformer-based registration network, because the SpikeReg recipe
requires a ReLU-only ANN of identical macro-architecture to the spiking
student; the question this paper asks is whether the registration
function learned by that ANN survives conversion to sparse spiking
computation, not whether the ANN itself sets a new accuracy ceiling.
For reference, the published TransMorph OASIS Dice is reported in the
range $\sim 0.78$--$0.80$ \citep{chen2022transmorph,hering2023learn2reg};
our teacher's $0.7480$ sits below that range, as expected for a stock
U-Net without long-range attention, but the relevant comparison for
SpikeReg is the within-architecture ANN-vs-SNN gap rather than the
absolute leaderboard position. A direct ANN-vs-SNN comparison under a
stronger (e.g.\ transformer) teacher is a separate experiment.
(v) The diffeomorphic SpikeReg-SVF variant has only been evaluated as a
single diagnostic run. It confirms that folding can be reduced, but the
large Dice loss means the present accuracy claims still apply only to the
direct-displacement model.

\paragraph{Future work.}
Beyond the limitations above, three directions look most promising.
\emph{First}, a topology-constrained SpikeReg should revisit the SVF
variant with loss weights, threshold calibration, and fine-tuning
schedules designed for integrated velocity fields rather than direct
displacement. \emph{Second}, deployment on a neuromorphic substrate
(Loihi~2 or SpiNNaker~2 SDK simulation, then hardware) would convert the
operation-count proxy into a measured-energy result. \emph{Third}, the
negative findings on KD and label-supervised teacher transfer suggest
that the ANN-to-SNN conversion literature needs a re-examination for
dense geometric prediction tasks beyond registration: any task whose output is a continuous, spatially structured field
(optical flow, depth estimation, surface normals, dense prediction in
remote sensing) falls in scope for the same family of analyses.

\paragraph{Significance.}
Despite these limitations, the central empirical signal is robust: a
spiking U-Net can perform dense 3D deformable brain MRI registration at
near-teacher Dice with an order-of-magnitude reduction in arithmetic
operations ($10.86\times$ fewer projected SynOps than measured ANN MACs per pair) and a
correspondingly larger projected arithmetic-energy reduction ($55.5\times$ under
the Horowitz $e_{\mathrm{AC}}/e_{\mathrm{MAC}}$ ratio, assuming event-driven execution), given a careful
ANN warm-start and per-layer threshold calibration. To our knowledge
this is the first demonstration of dense geometric regression under
sparse spiking computation in 3D medical imaging. It suggests that a
range of registration workloads (atlas propagation, longitudinal monitoring,
large-cohort morphometry) could in principle migrate toward energy-aware
neuromorphic compute without sacrificing registration accuracy.

\bibliographystyle{unsrtnat}
\bibliography{ref}  






\end{document}